%% file: main.tex
\documentclass[10pt,journal,compsoc,twocolumn]{IEEEtran}

\ifCLASSOPTIONcompsoc
  
  \usepackage[nocompress]{cite}
\else
  \usepackage{cite}
\fi
\ifCLASSINFOpdf
  \usepackage[pdftex]{graphicx}
  
\else
  
\fi
\usepackage{amsmath}
\usepackage{amssymb}
\usepackage{booktabs}
\usepackage{multirow}
\usepackage{xcolor}
\usepackage{subfig}
\usepackage{amsmath}

\input{macros}

\newcommand{\CS}[1]{\textcolor{black}{#1}}
\begin{document}
\title{Decoding Brain Representations by Multimodal Learning of Neural Activity and Visual Features}

\author{Simone~Palazzo,
	Concetto~Spampinato,~\IEEEmembership{Member,~IEEE,}
	Isaak~Kavasidis,
	Daniela~Giordano,~\IEEEmembership{Member,~IEEE,} Joseph Schmidt,
	and Mubarak~Shah,~\IEEEmembership{Fellow,~IEEE,}
	\IEEEcompsocitemizethanks{\IEEEcompsocthanksitem S. Palazzo, C. Spampinato, I. Kavasidis and D. Giordano are with the Department of Electrical, Electronic and Computer Engineering, University of Catania, Viale Andrea Doria, 6, Catania, 95125, Italy.\protect\\

E-mail: {palazzosim, cspampin, kavasidis, dgiordan}@dieei.unict.it
\IEEEcompsocthanksitem M. Shah and C. Spampinato are with the Center of Research in Computer Vision, University of Central Florida. E-mail: shah@crcv.ucf.edu
\IEEEcompsocthanksitem J. Schmidt is  with the Department of Psychology, University of Central Florida,  E-mail: Joseph.Schmidt@ucf.edu
}

}

\IEEEtitleabstractindextext{
\begin{abstract}

\change{This work presents a novel method of exploring human brain-visual representations, with a view towards replicating these processes in machines. The core idea is to learn plausible computational and biological representations by correlating human neural activity and natural images.
Thus, we first propose a model, \textit{EEG-ChannelNet}, to learn a brain manifold for EEG classification. After verifying that visual information can be extracted from EEG data, we introduce a multimodal approach that uses deep image and EEG encoders, trained in a siamese configuration, for learning a joint manifold that maximizes a compatibility measure between visual features and brain representations.\\
We then carry out image classification and saliency detection on the learned manifold.
Performance analyses show that our approach satisfactorily decodes visual information from neural signals. This, in turn, can be used to effectively supervise the training of deep learning models, as demonstrated by the high performance of image classification and saliency detection on out-of-training classes. The obtained results show that the learned brain-visual features lead to improved performance and simultaneously bring deep models more in line with cognitive neuroscience work related to visual perception and attention.
}

\end{abstract}

}

\maketitle

\IEEEdisplaynontitleabstractindextext

\IEEEpeerreviewmaketitle

\IEEEraisesectionheading{\section{Introduction}\label{sec:introduction}}
\input{introduction.tex}

\section{Related work}\label{sec:related}
\input{related.tex}

\section{Multimodal learning of visual-brain features}
\label{sec:joint}
\input{joint_new.tex}

\section{Image classification and saliency detection}\label{sec:saliency_method}
\input{saliency_method.tex}

\input{decoding_method.tex}

\section{Experiments and applications}\label{sec:experiments}
\input{intro_experiments.tex}

\subsection{Brain-Visual Dataset}\label{sec:dataset}
\change{
\input{dataset.tex}

}

\change{
\subsection{Model implementation}\label{sec:model_implement}
\input{eeg_encoder_impl.tex}
}

\subsection{\change{EEG classification}}
\label{sec:classification_eeg}

\CS{Our first experiment is intended to assess whether the architecture employed for the EEG encoder is able to extract visual-related information from  the EEG signals, and whether additional data post-processing (frequency filtering, temporal subsequencing) can affect performance. To this end, we trained the EEG encoder by itself to carry out visual classification from EEG signals, in the same fashion as ~\cite{Spampinato2016deep}. Note, this serves as a validation procedure of the EEG encoder architecture and as a baseline comparison with other state of the art models. This does not serve as any sort of pre-training for the full joint-embedding of the EEG and image encoders, rather to demonstrate the EEG carries visual information.}

To run this experiment, we append a softmax classification layer to the EEG-ChannelNet architecture (defined in Sect. \ref{sec:encoders} and shown in Figure \ref{fig:eeg_encoder}), and train the whole model to estimate the visual class corresponding to each EEG signal. The model is trained for 100 epochs, using a mini-batch size of 16, employing the Adam optimizer with suggested hyperparameters (learning rate: 0.001, $\beta_1$: 0.9, $\beta_2$: 0.999), and a mini-batch size of 16. As classification accuracy we report the accuracy obtained on the test set at the training epoch when the maximum accuracy on the validation set was achieved, according to the splits defined in Sect.~\ref{sec:dataset}.

\CS{Additionally, we also report the results obtained when training on certain frequency sub-bands and when employing only portions of the original signals.}

EEG signals contain several frequencies which are usually grouped into five bands: delta (0-4 Hz), theta (4-8 Hz), alpha (8-16 Hz), beta (16-32 Hz) and gamma (32 Hz to 95 Hz; this band can be further split into low- and high-frequency gamma with 55 Hz as cut-off). Gamma is believed to be responsible for cognitive processes, and high-frequency gamma (from 55 to 100 Hz) for short-term memory matching of recognized objects (working memory) and attention. Given that, our experimental design should elicit high gamma in participants. 
Thus, we also compute the performance when selecting only the above frequency bands: the results, given in Table \ref{tab:eeg_results_frequencies}, show that higher performance is achieved on high gamma frequencies, which is consistent with the cognitive neuroscience literature on attention, working memory and perceptual processes involved in visual tasks~\cite{CLAYTON2015188,TALLONBAUDRY1999151,JENSEN2007317}. Given these results, all following evaluations are carried out by applying a 55-95 Hz band-pass filter.

\begin{table}[]
    \change{
    \centering
    \begin{tabular}{ccc}
    \toprule
    {\textbf{Band}} & {\textbf{Frequency - Hz}} & \textbf{Accuracy}\\
    \midrule
    Theta, alpha and beta & 5-32  & 19.7\%\\
    Low frequency gamma   & 32-45 & 26.1\%\\
    \textbf{High frequency gamma}  & \textbf{55-95 }& \textbf{48.1\%}\\
    All gamma  & 32-95 & 40.0\%\\
    All frequencies       & 5-95  & 31.3\%\\
    \bottomrule
    \end{tabular}
    }
    \caption{\change{EEG classification accuracy using different EEG frequency bands. We discard the 45--55 frequencies because of the power line frequency at 50Hz. }}
    \label{tab:eeg_results_frequencies}
\end{table}

We then evaluate performance when using the entire time course of the EEG signals and a set of temporal EEG sub-sequences (of size 220 ms, 330 ms, or 440 ms) to understand when the neural response results in the strongest classification performance. Note that the EEG encoder architecture described in Sect.~\ref{sec:model_implement} was not adapted for this evaluation: the convolutional portion of the model is unaffected by the reduced temporal length, which reflects only the input size to the fully-connected layer. Results are shown in Tab. \ref{tab:eeg_results}, and show that when using shorter EEG segments, performance is lower than when we use the entire time course (20-460 ms). Additionally, leaving out the first 110 ms negatively affects the performance, suggesting that human perceptual processes tied to stimulus onset are critical for decoding image classes. Furthermore, adding the last 110 ms increases the performance by 5 percentage points (20-350 ms vs 20-460 ms), meaning that the later, more cognitive operations performs a further refinement of the learned features which enhances classification; this is also in line with the neurocognitive literature about experiments based on event-related potentials (ERP)~\cite{luck2014introduction}. The almost equal performance between 20-240 ms (39.4\%) and 240-460 ms (38.9\%) may suggest a balanced importance between an initial low-level visual feature extraction and a more cognitive aggregation into different abstraction levels (evidence of this processing can be found in the saliency detection analysis in Sect. \ref{sec:saliency}).

\begin{table}[]
\change{
    \centering
    \begin{tabular}{cc}
    \toprule
    {\textbf{EEG Time interval [ms]}} & \textbf{EEG Classification Accuracy}\\
    \midrule
    20-240  & 39.4\%\\
    20-350  & 43.8\%\\
    \textbf{20-460}  & \textbf{48.1\%}\\
    130-350 & 23.5\%\\
    130-460 & 26.3\%\\
    240-460 & 38.9\%\\
    \midrule
    & \\
    \midrule
    \multicolumn{2}{c}{\textbf{SOTA methods}}\\
    \midrule
    {Method} & \textbf{EEG Classification Accuracy} \\
    \midrule
    \cite{Spampinato2016deep} & 21.8\%\\
	\cite{Lawhern_2018} & 31.9\%\\
    \cite{NIPS2017_7048} & 31.7\% \\
    \bottomrule
    \end{tabular}
}
    \caption{\change{(Top) EEG classification accuracy using different EEG time intervals with data filtered in the [55-95] Hz band. (Bottom): Classification performance of state of the art methods using the whole EEG time course (i.e., 20-460 ms).}}
    \label{tab:eeg_results}
\end{table}

Finally, we compare classification performance achieved by our EEG encoder and other state-of-the-art methods, namely \cite{Spampinato2016deep}\footnote{\CS{Note that its performance are lower than those  reported in  \cite{Spampinato2016deep}, since in that work, frequency filtering was carried out incorrectly, i.e., DC component was not removed, leaving the EEG drift that induced a bias in signals and, consequently, higher performance.}} \cite{Lawhern_2018},\cite{NIPS2017_7048}, using high-frequency gamma band data, i.e., 55-95 Hz. EEG classification accuracy on the test split is given in Tab.~\ref{tab:eeg_results} and shows that our approach reaches an average classification accuracy of 48.1\%, outperforming previous methods, such as EEGNet, which, only achieves a maximum accuracy of 31.9\%.

\subsection{Siamese network training for classification}\label{sec:classification}
\input{classification.tex}

\subsection{Saliency detection}\label{sec:saliency}
\input{saliency.tex}

\subsection{Decoding Brain Representations}\label{sec:decoding_experiments}

\input{eeg_analysis.tex}

\input{brain_representations.tex}

\section{Conclusion}\label{sec:conclusions}
\input{conclusions.tex}

\ifCLASSOPTIONcompsoc
  \section*{Acknowledgments}
\else
  \section*{Acknowledgment}
\fi
The authors would like to thank Dr. Martina Platania for supporting the data acquisition phase, Dr. Demian Faraci for the experimental results, and NVIDIA for the generous donation of two Titan X GPUs.

\ifCLASSOPTIONcaptionsoff
  \newpage
\fi

\bibliographystyle{IEEEtran}

\bibliography{main}

\begin{IEEEbiography}[{\includegraphics[width=1in,height=1.25in,clip,keepaspectratio]{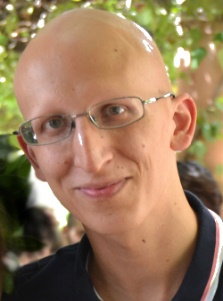}}]{Simone Palazzo} is an assistant professor at the University of Catania, Italy. His research interests are in the areas of deep learning, computer vision, integration of human feedback into AI systems, medical image analysis. He has been part of the program committees of several workshops and conferences on computer vision. He has co-authored over 50 papers in international refereed journals and conference proceedings.
\end{IEEEbiography} \vspace{-1.5cm}

\begin{IEEEbiography}[{\includegraphics[width=1in,height=1.25in,clip,keepaspectratio]{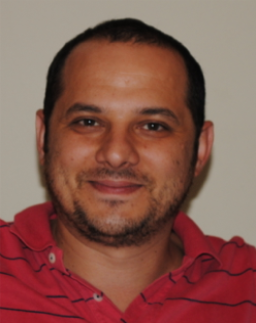}}]{Concetto Spampinato} is an assistant professor at the University of Catania, Italy. He is also Courtesy Faculty member of the Center for Research in Computer Vision at the Unviersity of Central Florida (USA). His research interests lie mainly in the artificial intelligence, computer vision and pattern recognition research fields with a particular focus on human-based and brain-driven computation systems. He has co-authored over 150 publications in international refereed journals and conference proceedings.
\end{IEEEbiography} \vspace{-1.5cm}

\begin{IEEEbiography}[{\includegraphics[width=1in,height=1.25in,clip,keepaspectratio]{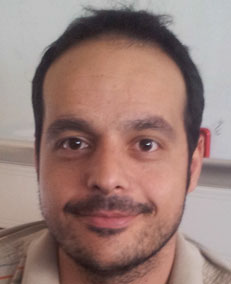}}]{Isaak Kavasidis} is an assistant researcher at the University of Catania in Italy. His research interests include the areas of medical data processing and brain data processing using machine and deep learning methods and the decoding of human brain functions and transfer to computerized methods. In 2014, he participated in the Marie Curie RELATE ITN project as an experienced researcher. He has co-authored more than 40 scientific papers in peer-reviewed international conferences and journals
\end{IEEEbiography} \vspace{-1.5cm}

\begin{IEEEbiography}[{\includegraphics[width=1in,height=1.25in,clip,keepaspectratio]{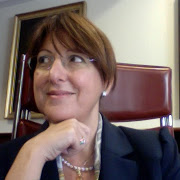}}]{Daniela Giordano} is an associate professor at the University of Catania, Italy. She also holds the Ph.D. degree in Educational Technology from Concordia University, Montreal (1998). Her main research interests include advanced learning technology, knowledge discovery, and information technology in medicine. She has co-authored over 200 publications in international refereed journals and conference proceedings.
\end{IEEEbiography} \vspace{-1.5cm}

\begin{IEEEbiography}[{\includegraphics[width=1in,height=1.25in,clip,keepaspectratio]{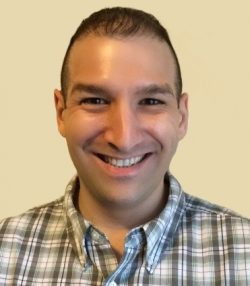}}]{Joseph Schmidt} is an assistant professor at the University of Central Florida, Department of Psychology and he currently oversees the Attention and Memory Lab. Dr. Schmidt’s research investigates how memory representations of target objects affect deployments of attention. He has developed expertise in many psychophysiological techniques including eye tracking and electroencephalogram (EEG)/event-related potentials (ERPs). He has co-authored about many publications in international refereed journals.
\end{IEEEbiography}  \vspace{-1.5cm}

\begin{IEEEbiography}[{\includegraphics[width=1in,height=1.25in,clip,keepaspectratio]{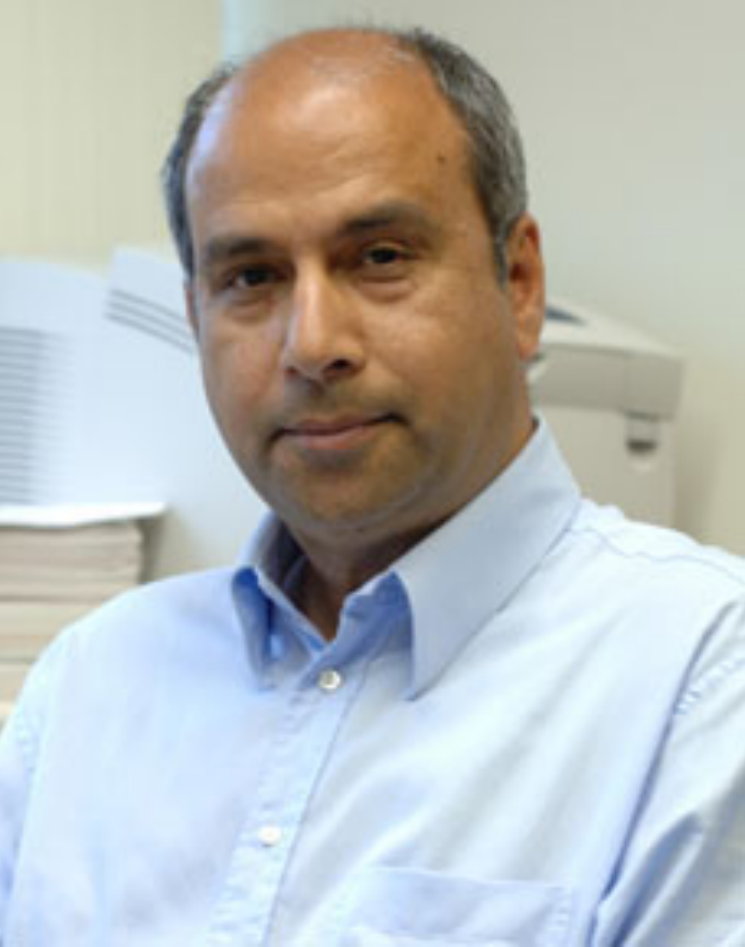}}]{Mubarak Shah} is the trustee chair professor of computer  science and the  founding  director  of the Center for Research in Computer Vision at University of Central Florida. His research interests include video surveillance, visual tracking, human activity recognition, visual analysis of crowded  scenes,  video  registration,  UAV  video analysis, and so on. He is a fellow of the IEEE, AAAS, IAPR, and SPIE.
\end{IEEEbiography}
\end{document}

%% file: macros.tex
\newcommand{\mytilde}{\raise.17ex\hbox{$\scriptstyle\mathtt{\sim}$}}

\newcommand{\change}[1]{\textcolor{black}{#1}}

%% file: introduction.tex
\IEEEPARstart{H}{uman} visual capabilities are nearly rivaled by artificial systems, mainly thanks to the  recent advances in deep learning. Indeed, deep feedforward and recurrent neural networks, loosely inspired by the primate visual system, have led to a significant boost in performance of computer vision, natural language processing, speech recognition and game playing. In addition to the significant performance gain in such tasks, the representations learned by deep computational models appear to be highly correlated with brain representations. For example, correlations can be found between brain representations in the visual pathway and the hierarchical structures of layers in deep neural networks (DNNs)~\cite{pmid28530228,pmid27282108}. 
These findings have paved the way for multidisciplinary efforts involving cognitive neuroscientists and artificial intelligence researchers, with the aim of reverse-engineering the human mind and its adaptive capabilities \cite{Spampinato2016deep, 8237631,pmid21945275,pmid23932491}. 
Nevertheless, this multidisciplinary  field is still in its infancy. Indeed, the existing computational neural models loosely emulate computations and connections of biological neurons, but they often ignore feedforward and feedback neural interactions. For example, visual recognition in humans appears to be mitigated by a multi-level aggregation of information being processed forward and backward across cortical brain regions \cite{pmid11690606,pmid21438683,pmid21415848,pmid22325196}. Recent approaches\cite{pmlr-v80-wen18a}, inspired by the  hierarchical predictive coding in neuroscience \cite{pmid23663408,pmid23177956}, have attempted to encode such additional information into computational models by proposing recurrent neural networks with feedforward, feedback, and recurrent connections. \CS{These models have shown promising performance in visual classification tasks and demonstrate that understanding the human brain in more detail may allow us to transfer that knowledge to engineering models to create better machines. 
Clearly, before human level classification performance can be transferred to computational models, 
it is first necessary to better understand the human visual system. To accomplish this, we intend to correlate neural activity data recorded from human subjects while performing specific tasks with our computational models developed to complete the same task}. By investigating the learned computational representations and how they correlate with neural activity over time, it is possible to infer, analyze and eventually replicate complex brain processes in machines. \\
In this paper, we first propose a model to learn neural representations by classifying brain responses to natural images. Then, we  introduce a multimodal approach based on deep learning EEG and image encoders, trained in a siamese configuration. Given EEG brain activity data from several human subjects performing visual categorization of still images, our multimodal approach learns a joint brain-visual embedding and finds similarities between brain representations and visual features.
The embedding is then used to perform image classification, saliency detection, and to hypothesize  possible representations generated in the human brain for visual scene analysis. \\
In particular, this paper demonstrates that a) neural activity data can be used to provide richer supervision to  deep learning models, resulting in visual classification and saliency detection methods aligned with human neural data; b) \CS{joint artificial intelligence and cognitive neuroscience efforts may lead to uncover neural processes involved in human visual perception by  maximizing the similarity of deep models with human neural responses. Indeed, we propose a method to extract visual saliency (and its evolution over time), as well as to localize the cortical region producing such information; and  c) there is potential similarity between computational representations and brain processes, providing interesting insights about consistency between biological and deep learning models. }\\
Summarizing, the proposed approach to learn and correlate brain processes to visual cues, both in time and space, results in a twofold contribution:
\CS{
\begin{itemize}
    \item \textbf{Artificial Intelligence.} We introduce new models for decoding EEG signals related to visual tasks with state-of-the-art performance and in a biologically-plausible way. Moreover, our approach allows to automatically identify computational features that consistently match human neural activity, representing a new direction to help explain AI models.
    \item \textbf{Cognitive Neuroscience.} Our approach is a step forward towards providing  cognitive neuroscientists  with AI-based methodology for understanding neural responses both in space and time, without the need to design experiments with multiple subjects and trials. When highly accurate AI is designed, it will allow cognitive neuroscientists to simulate human responses rather than collect significant amounts of costly data.
\end{itemize}}
\vspace{0.5cm}

The paper is organized as follows. In the next section, we review recent methods combining brain data and computational models, as well as related approaches to multimodal learning. In Sect.~\ref{sec:joint} we describe the core of our approach and review the specific framework used to learn a joint brain-visual embedding. Next we describe the methods to extract the most relevant visual information from the image and the brain activity patterns, as well as their interpretation (Sect.~\ref{sec:saliency_method}--\ref{sec:decoding}). Sect.~\ref{sec:experiments} reports the achieved experimental results related to image classification and saliency detection, and shows the learned representations that mostly affect visual categorization. In the last section, the conclusion and future directions are presented.

%% file: related.tex
Our work most directly relates to the fields of EEG data classification, computational neuroscience for brain decoding, machine learning guided by brain activity and multimodal learning. The recent state of the art in these areas are briefly reviewed in this section. \\

\change{
\noindent {\bf EEG data classification}. In recent years, the deep learning--based approach to classification EEG data has grown in popularity (a comprehensive review can be found in \cite{roy_review_eeg}). Most of these methods propose a custom AI solutions used to categorize data in a BCI (Brain Computer Interface) application (e.g., motor imagery, speech imagery, emotion recognition, etc.)~\cite{AAAI1816107,2607-18}, in clinical applications~\cite{YanS19.004,7727334} (e.g., epilepsy detection and prediction) or for monitoring cognitive functions~\cite{8275511,TANG201711} (mental workload, engagement, fatigue, etc.). The above work focuses on the classification of a few categories (from binary classification to less than 10 classes), and none of them have the primary objective of understanding human visual processes (in space and time). Furthermore, most of the proposed models have been applied to single BCI paradigms, with very controlled stimuli. \CS{The obvious concern is that these methods may fail to generalize and their capabilities may collapse with small changes to the stimuli and task. 
EEGNet attempts to address this generalization problem \cite{Lawhern_2018}. To do so, the authors implement a compact convolutional neural network for EEG classification. The model employs depth-wise separable convolutions and can be applied to different EEG experiments, leading to good performance in several tasks: P300 visual-evoked potentials, error-related negativity responses (ERN), movement-related cortical potentials (MRCP), and sensory motor rhythms (SMR). Analogously, \cite{NIPS2017_7048} proposes parameterized convolutional filters that learn relevant information at different frequency bands while targeting synchrony. }\\
The EEG classification approach proposed in this paper aims to improve the architectural design concepts of \cite{Lawhern_2018,NIPS2017_7048} by modelling more general spatio-temporal features of neural responses with a goal of  supporting cognitive neuroscience studies to improve the interpretability of human neural data in time and space.}\\

\noindent {\bf Computational neuroscience for decoding brain representations}. Decoding brain representations has been a long sought objective and it still is a great challenge of our times.
In particular, cognitive neuroscience works have made great progress in understanding neural representations originating in the primary visual cortex (V1).  Indeed, it is known that the primary visual cortex is a retinotopically organized series of oriented edge and color detectors\cite{pmid25681421} that feed-forward into neural regions focused on more complex shapes and feature dimensions, which operate over larger receptive fields in areas V4 \cite{pmid26053241}, before finally arriving at object and category representations in the inferior temporal (IT) cortex \cite{pmid16272124}. 
Neuroimaging methods, such as fMRI, MEG, and EEG, have been crucial for these findings. However, to recreate human level neural representations that fully represent our visual processes would require precisely monitoring the activity of every neuron in the brain simultaneously. Although these methods are clearly incapable of accomplishing this lofty goal, they contain enough information to accurately reconstruct many visual experiences \cite{pmid29176609}. To that end, brain representation decoding has recently examined the correlation between neural activity data and computational models \cite{pmid27282108,pmid28530228}.
However, these approaches mainly perform simple correlations between deep learned representations and neuroimaging data and, according to the obtained outcomes, draw conclusions about brain representations, which is too simplistic from our point of view. Indeed, the core point of our idea is that understanding the human visual system will come as a result of training automated models to maximize signal correlation between brain activity and the evoking stimuli, not as a pure analysis of brain activity data. 
In addition, while most of the methods attempt to decode brain representations using brain images from high spatial resolution fMRI, our work is the first one to employ EEG data that, despite being lower spatial resolution, has higher temporal resolution, which makes it more suitable to decode fast brain processes like those involved in the visual pathway. Additionally, unlike fMRI, EEG is portable, ambulatory, and can even be used wirelessly, traits that would improve any BCI.\\

\noindent {\bf Machine learning guided by brain activity}. The intersection and overlap between machine learning and cognitive neuroscience has increased significantly in recent years.

Deep learning methods are used, for instance, for neural response prediction\cite{NIPS2013_4991,pmid24812127,pmid19104670}, and, in turn, biologically-inspired mechanisms such as coding theory\cite{pmlr-v80-wen18a}, working memory\cite{pmid27732574} and attention \cite{pmlr-v37-gregor15,pmlr-v37-xuc15} are increasingly being adopted.
However, to date, human cognitive abilities still seem too complex to be understood computationally, and a data-driven approach for ``reverse-engineering'' the human mind might be the best way to inform and advance artificial intelligence \cite{pmid27881212}. Under this scenario, recent studies have employed neural activity data to constrain model training. For example, in our recent work~\cite{Spampinato2016deep}, we mapped visual features learned by a deep feed-forward model to brain-features learned directly from EEG data to perform automated visual classification. The authors of \cite{pmid29599461} employed fMRI data to bias the output of a machine learning algorithm and push it to exploit representations found in visual cortex. This work resembles one of the first methods relying on brain activity data to perform visual categorization \cite{4587618}, with the distinction that the former, i.e., \cite{pmid29599461}, explicitly utilizes neural activity to weigh the training process (similarly to \cite{pmid26353347}), while the latter, i.e. \cite{4587618}, proposes a kernel alignment algorithm to fuse the decision of a visual classifier with brain data.\\ 
In this paper, we propose a deeper interconnection between the two fields: instead of using neural data as a signal to weigh computationally-learned representations, we learn a mapping between images and the corresponding neural activity, so that visual patterns are related in a one-to-one fashion to neural processes. This mapping, as we demonstrate in the experimental results, may reveal much more information about brain representations and be able to guide the training process in a more intrinsic and comprehensive way. Thus, our approach is not just a hybrid machine learning method that is inspired or constrained by neural data, but a method that implicitly finds similarities between computational representations, visual patterns and brain representations, and uses them to perform visual tasks.\\
\\

{\bf Multimodal learning}. Our research utilizes multimodal learning. We exploit the fact that real-world information comes from multiple modalities, each carrying different --- yet equally useful --- content for building intelligent systems. 
Multimodal learning methods \cite{conf/icml/NgiamKKNLN11,NIPS2014_5279,JMLR:v15:srivastava14b}, in particular,  attempt to learn embeddings by finding a joint representation that encodes the real-world features of the stimulus  across multiple modalities to create a common concept of the input data.

An effective joint representation must preserve both intra-modality similarity (e.g., two similar images should have close vector representations in the joint space; likewise, two equivalent text descriptions should have similar representations as well) and inter-modality similarity (e.g., an image and a piece of text describing the content of that image should be closer in the joint space than an image and an unrelated piece of text).
Following this property, most methods find correspondences between visual data and text \cite{Venugopalan_2017_CVPR,NIPS2017_6658,5540120,JMLR:v15:srivastava14b} or audio \cite{zhao2018sound,10.1007/978,Aytar:20166,7952132} to support either discriminative tasks (e.g., classification) or prediction of one modality conditioned on another (e.g., image synthesis or retrieval).  
For the former type of methods, captions and tags have been used to improve accuracy of both shallow and deep classifiers \cite{5540120,Huiskes:2010}. Analogously, \cite{10.1007/978} used audio to supervise visual representations; \cite{Aytar:20166,7952132} used vision to supervise audio representations; \cite{Arandjelovic_2017_ICCV} used sound and vision to jointly supervise each other; and \cite{zhao2018sound} investigated how to separate and localize multiple sounds in videos by analyzing motion and semantic cues. 
Other works, have instead focused on predicting missing data in one modality from the other modality, for example, generate text descriptions from images and vice versa \cite{43274,KarpathyF17,DonahueHGRVDS15,pmlr-v48-reed16,MansimovPBS15}. Reed et al. in \cite{pmlr-v48-reed16} propose a joint representation space to condition generative adversarial networks (GANs) for synthesizing images from text descriptions. Similarly, Mansimov et al. \cite{MansimovPBS15} synthesized images from text captions using a variational autoencoder. In our recent paper \cite{Kavasidis:2017}, we used an embedding learned from brain signals to synthesize images both using GANs and variational autoencoders in a brain-to-image effort.

In this paper, our approach is inspired by the methods that learn a shared multimodal representation, with several crucial differences. 
First, one of the modalities we utilize is brain activity data (EEG), which is almost certainly noisier than then text/audio. This makes it much harder to discover relationships between the visual and brain modalities. In this sense, our approach is intended to improve prediction accuracy and to act as a knowledge discovery tool to uncover brain processes. Thus, our main objective is to learn a reliable joint representation and explore the learned space to find correspondences between visual and brain features that can uncover brain representations; these, in turn, can be employed to build better deep learning models. \\
In addition, the proposed deep multimodal network, consisting of two encoders (one per modality), is trained in a siamese configuration and employs a loss function enforcing the learned embedding to be representative of intra-class differences between samples, and not just of the inter-class discriminative features (as done, for instance, in \cite{pmlr-v48-reed16}).

%% file: joint_new.tex
Neural activity (recorded by EEG) and visual data have very different structures, and finding a common representation is not trivial. Previous approaches \cite{Spampinato2016deep} have attempted to find such representations by training individual models: for example, by first learning brain representations by training a recurrent classifier on EEG signals, and then training a CNN to regress the visual features to brain features for corresponding EEG/image pairs. While this provides useful representations, the utility of the learned features is strongly tied to the proxy task employed to compute the initial representation (e.g., image classification), and focuses more on learning class-discriminative features than on finding relations between EEG and visual patterns. 

\begin{figure}
	\centering
	\includegraphics[width=0.5\textwidth]{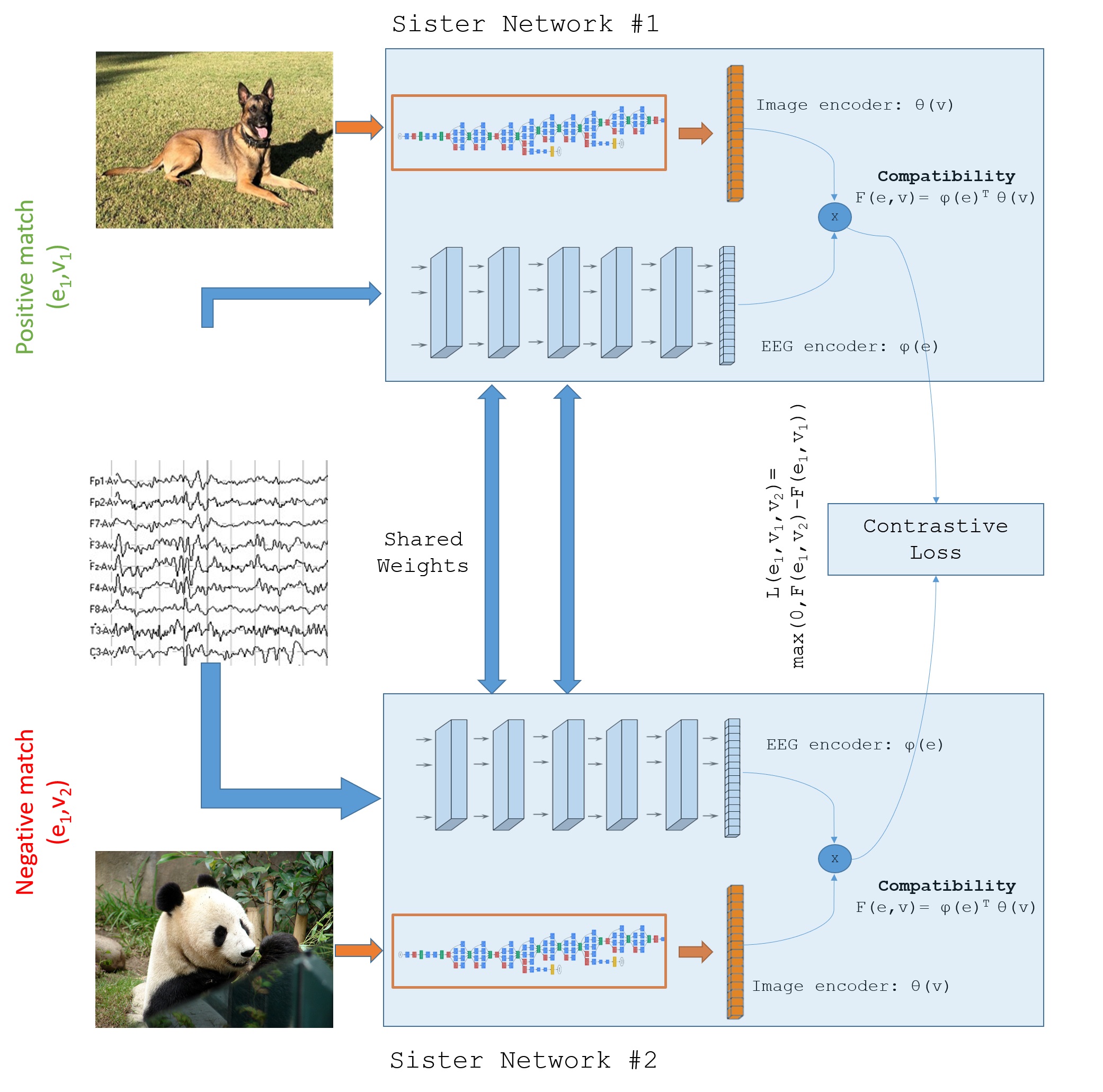}
	\caption{\textbf{Siamese network for learning a joint brain-image representation}. The idea is to learn a space by maximizing a compatibility function between two embeddings of each input representation. Given a positive match between an image and the related EEG from one subject, and a negative match between the same EEG and a different image, the network is trained to ensure a closer similarity (higher compatibility) between related EEG/image pairs than unrelated ones.}
	\label{fig:joint}
\end{figure}

Hence, we argue that any transformations from human neural signals and images to a common space should be learned jointly by maximizing the similarity between the embeddings of each input representation. To this aim, we define a siamese network for learning a structured joint embedding between EEG signals and images using deep encoders, and maximize a measure of similarity between the two modalities. The architecture of our model is shown in Fig.~\ref{fig:joint}.

More formally, let $\mathcal{D} = \{ e_i, v_i\}_{i=1}^N$ be a dataset of neural signal samples and images, such that each neural (EEG) sample~$e_i$ is recorded on a human subject in response to viewing image~$v_i$. Ideally, latent information content should be shared by $e_i$ and $v_i$. Also, let $\mathcal{E}$ be the space of EEG signal samples and $\mathcal{V}$ the space of images. The objective of our method is to train two encoders that respectively map neural responses and images to a common space $\mathcal{J}$, namely $\varphi: \mathcal{E} \rightarrow \mathcal{J}$ and $\theta: \mathcal{V} \rightarrow \mathcal{J}$.

In other approaches for structured learning (e.g.~\cite{pmlr-v48-reed16}), the training of the encoders is proxied as a classification problem based on the definition of a \textit{compatibility function} $F: \mathcal{E} \times \mathcal{V} \rightarrow \mathbb{R}$, that computes a similarity measure as the dot product between the respective embeddings of an EEG/image pair:
\begin{equation}
 F(e,v) = \varphi(e)^T \theta(v)~.
 \label{eq:comp}
\end{equation}

While we employ the same modeling framework, we formulate the problem as an embedding task whose only objective is to maximize similarity between corresponding pairs, without implicitly performing classification, as this would take us back to the limitation of \cite{Spampinato2016deep}, i.e., learning representations tied to the classification task.\\
In order to abstract the learning process from any specific task, we train our siamese network with a triplet loss aimed at mapping the representations of matching EEGs and images to nearby points in the joint space, while pushing apart mismatched representations. We can then stick to the structured formulation of the compatibility function in Eq.~\ref{eq:comp} by employing $F$ directly for triplet loss computation. Thus, given two pairs of EEG/image $(e_1, v_1)$ and $(e_2, v_2)$, we consider $e_1$ as the \emph{anchor} item, $v_1$ as the \emph{positive} item and $v_2$ as the \emph{negative} item. Using compatibility $F$ (which is a similarity measure rather than a distance metric, as is more commonly used in triplet loss formulations), the loss function employed to train the encoders becomes:
\begin{equation}
 L(e_1, v_1, v_2) = \max \{ 0, F(e_1, v_2) - F(e_1, v_1) \}~.
 \label{eq:loss}
\end{equation}
This equation assigns a zero loss only when compatibility is larger for $(e_1,v_1)$ than for $(e_1,v_2)$. Note that class labels are not used anywhere in the equation. This makes sure that the resulting embedding does not just associate class-discriminative vectors to EEG and images, but tries to extract more comprehensive patterns that explain the relations between the two data modalities. Also, there is no margin term in Eq.~\ref{eq:loss}, as would be typical in hinge loss formulations of a triplet loss. This is due to $(e_1,v_1)$ and $(e_2,v_2)$ possibly being members of the same visual class, and forcing a minimum distance between the same-class items is not strictly needed: as long as the learned representation assigns a larger compatibility to matching EEG/image pairs and learns general and meaningful patterns, class separability would still be implicitly achieved.\\
In the next subsection we present the architectures of the EEG encoder $\varphi$ and  the image encoder $\theta$.

\subsection{\change{Encoders' architectures}}
\label{sec:encoders}

\begin{figure}
 \centering
 \includegraphics[width=0.50\textwidth]{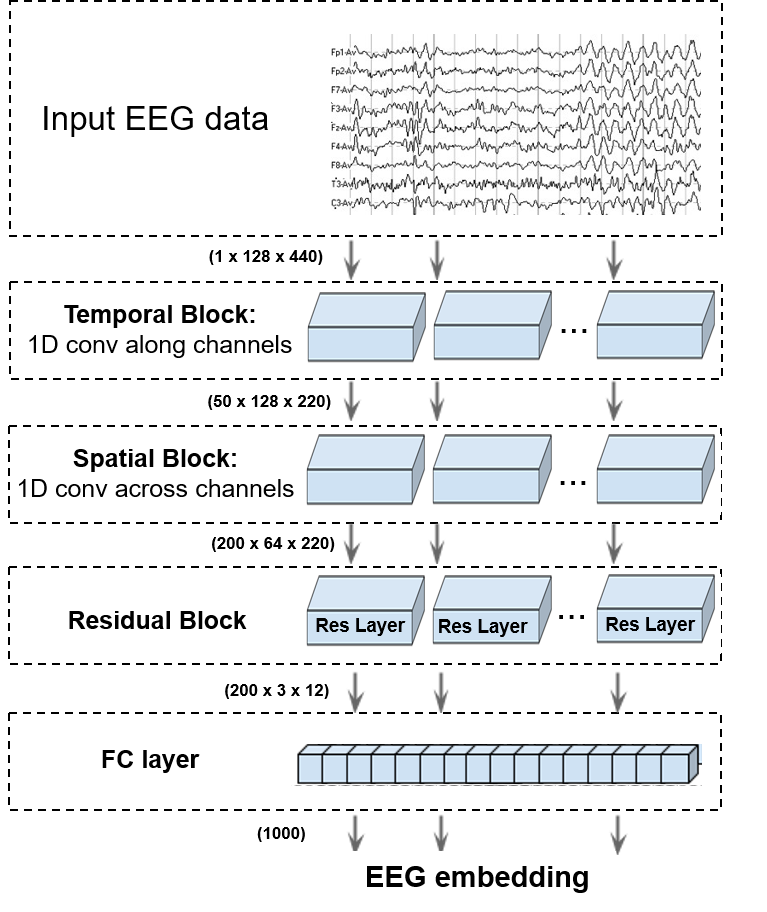}\\ 
  \includegraphics[width=0.50\textwidth]{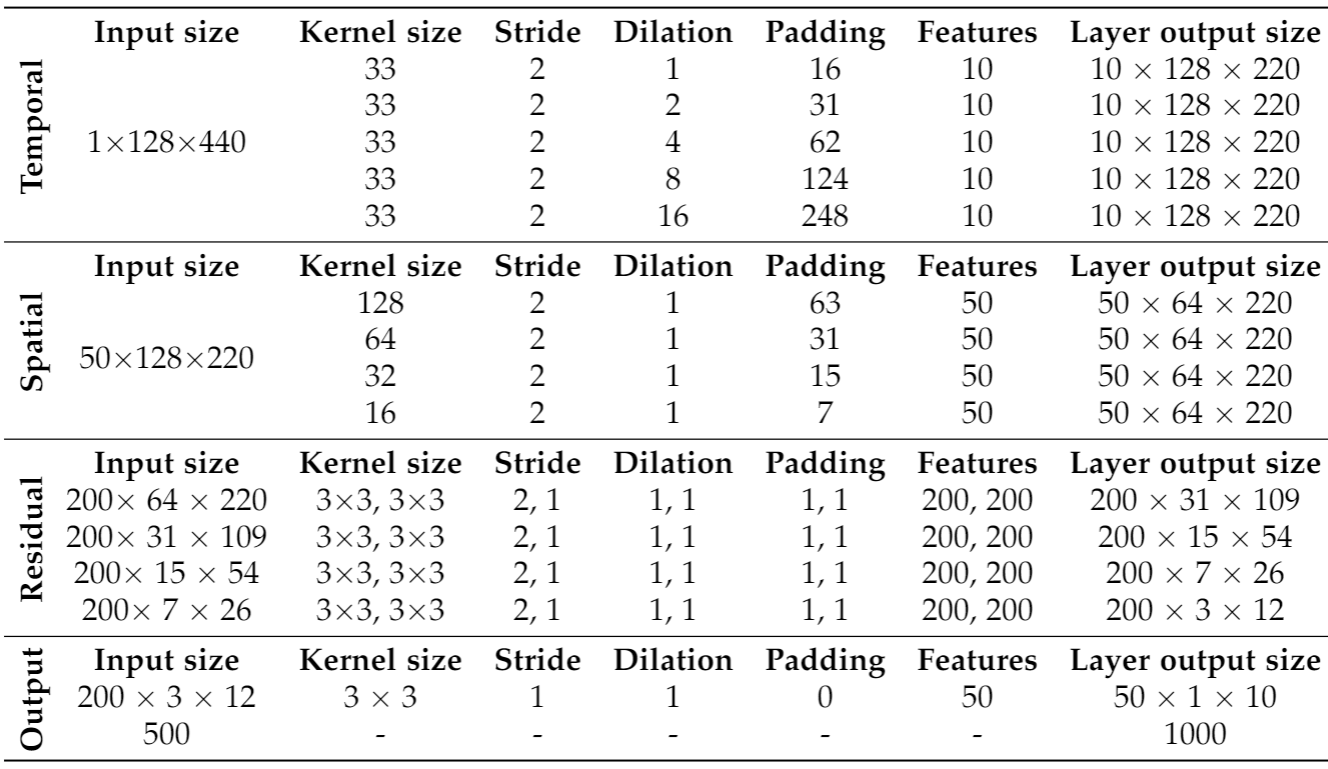}
 \caption{\change{\textbf{Detailed EEG-ChannelNet Architecture}. The EEG signal is first processed by a bank of concatenated 1D convolutions over channels (\emph{temporal block}), followed by a bank of concatenated 1D convolutions across channels (\emph{spatial block}). The resulting features are then processed by a cascade of residual layers, followed by a final convolution and a fully-connected layer projecting to the joint embedding dimensionality.}}
 \label{fig:eeg_encoder}
\end{figure}

\change{
EEG encoder $\varphi(\cdot)$, which maps neural activity signals to the joint space $\mathcal{J}$, is a convolutional network, dubbed \emph{EEG-ChannelNet} and shown in Fig. \ref{fig:eeg_encoder}, with a \emph{temporal block}, a \emph{spatial block} and a \emph{residual block}, which process different dimensions of the input signal in different steps following a hierarchical approach.} 

\change{The \emph{temporal block} first processes the input signal along the temporal dimension, by applying 1D convolutions to each channel independently, with the twofold purpose of extracting significant features and reducing the size of the input signal. In order to be able to capture patterns at multiple temporal scales, the temporal block internally includes a set of 1D convolutions with different kernel dilation values~\cite{YuK15}, whose output maps are then concatenated. The role of the temporal block is to extract information representing significant temporal patterns \emph{within} each channel.}

\change{The following \emph{spatial block} aims instead at finding correlations between different channels at corresponding time intervals, by applying 1D convolutions over the channel dimension. 
To clarify this aspect, note that an input EEG signal of size $C\times L$ (with $C$ being the number of channels, and $L$ the temporal length) will be transformed by the temporal block into a tensor of size $F\times C \times L_T$, with $F$ being the number of concatenated feature maps and $L_T$ being the ``new'' temporal dimension, after the application of the 1D convolution. Each element of this tensor will not temporally correspond to a single sample in the original signal, but it will ``cover'' a specific temporal receptive field, depending on the kernel size and dilation.}
\change {The spatial block, then, operates on the feature and channel dimensions across each element in the $L_T$ dimension, with the objective of analyzing spatial correlations at corresponding times (on multiple scales). Similar to the temporal block, the spatial block also consists of multiple 1D convolutional layers whose outputs are concatenated. In this case, the channel dimension is sorted so that ``rows'' of channels (according to the 10-20 layout depicted in Fig.~\ref{fig:eeg_cap}) are appended consecutively in the signal matrix; then, each spatial 1D convolution operates with different kernel sizes. 
All convolutional layers in the temporal and spatial blocks are followed by batch normalization and ReLU activations.
Once the model has worked independently on the temporal and spatial dimensions, the final \emph{residual block}, consisting of a set of residual layers~\cite{He2015deep}, performs 2D convolution on the spatio-temporal representation to find more complex relations and representations from the signal. Each residual layer performs two convolutions (with batch normalization and ReLU activation) before summing the input to the residual. The output is then provided to a final convolutional layer followed by a fully-connected layer, having the same size as the joint embedding dimensionality.}

\change {The proposed encoder is first tested for EEG classification, by suitably adding a softmax layer after the fully connected one, in order to understand its capabilities to decode visual information from neural data (see Sect. \ref{sec:classification_eeg}). Afterwards, the encoder is trained using the siamese schema presented earlier.} 

Visual encoder, $\theta(\cdot)$ maps, instead, images to the joint space $\mathcal{J}$ through convolutional neural networks. We use a pre-trained CNN to extract visual features and feed them to a linear layer for mapping to the joint embedding space. 
Differently from \cite{pmlr-v48-reed16}, we learn the compatibility function in an end-to-end fashion, also by fine-tuning the image encoder, in order to better identify low- and middle- level visual-brain representations, which --- suitably decoded --- may provide hints on what information is used by humans when analyzing visual scenes.

\change{}{}

%% file: saliency_method.tex
Our siamese network learns visual and EEG embeddings in order to maximize the similarities between  images and related neural activities. We can leverage the learned manifold for performing visual tasks. 
In cognitive neuroscience there is converging evidence that: a)  brain activity recordings contain information about visual object categories (as also demonstrated in \cite{Spampinato2016deep}) and b) attention influences the processing of visual information even in the earliest areas of the primate visual cortex \cite{TREUE2003428}. In particular, bottom-up sensory information and top-down attention mechanisms seem to fuse in an integrated saliency map, which in turn, distributes across the visual cortex. 
Thus, EEG recordings in response to visual stimuli should encode both visual class and saliency information. However, for image classification we can simply use the trained encoders as feature extractors for a subsequent classification layer (see performance evaluation in Sect.~\ref{sec:experiments}), whereas for saliency detection we designed a \textit{multiscale suppression-based} approach, inspired by the methods identifying pixels relevant to CNN neuron activations (e.g., \cite{zeiler2014visualizing}), that analyzes fluctuations in the compatibility measure $F$ (\ref{eq:comp}). The idea is based on measuring how brain-visual compatibility varies as image patches are suppressed at multiple scales. Indeed, the most important features in an image are those that, when inhibited in an image, lead to the largest drop in compatibility score (computed by feeding an EEG/image pair to the siamese network proposed in the previous section) with respect to the corresponding neural activity signal. Thus, we employ compatibility variations at multiple scales for \emph{saliency detection}. Note that, for this approach to work, the EEG encoder must have learned to identify patterns related to specific visual features in the observed image, so that the absence of those features reflects on smaller similarity scores on the joint embedding space.

\begin{figure}
	\centering
	\includegraphics[width=0.5\textwidth]{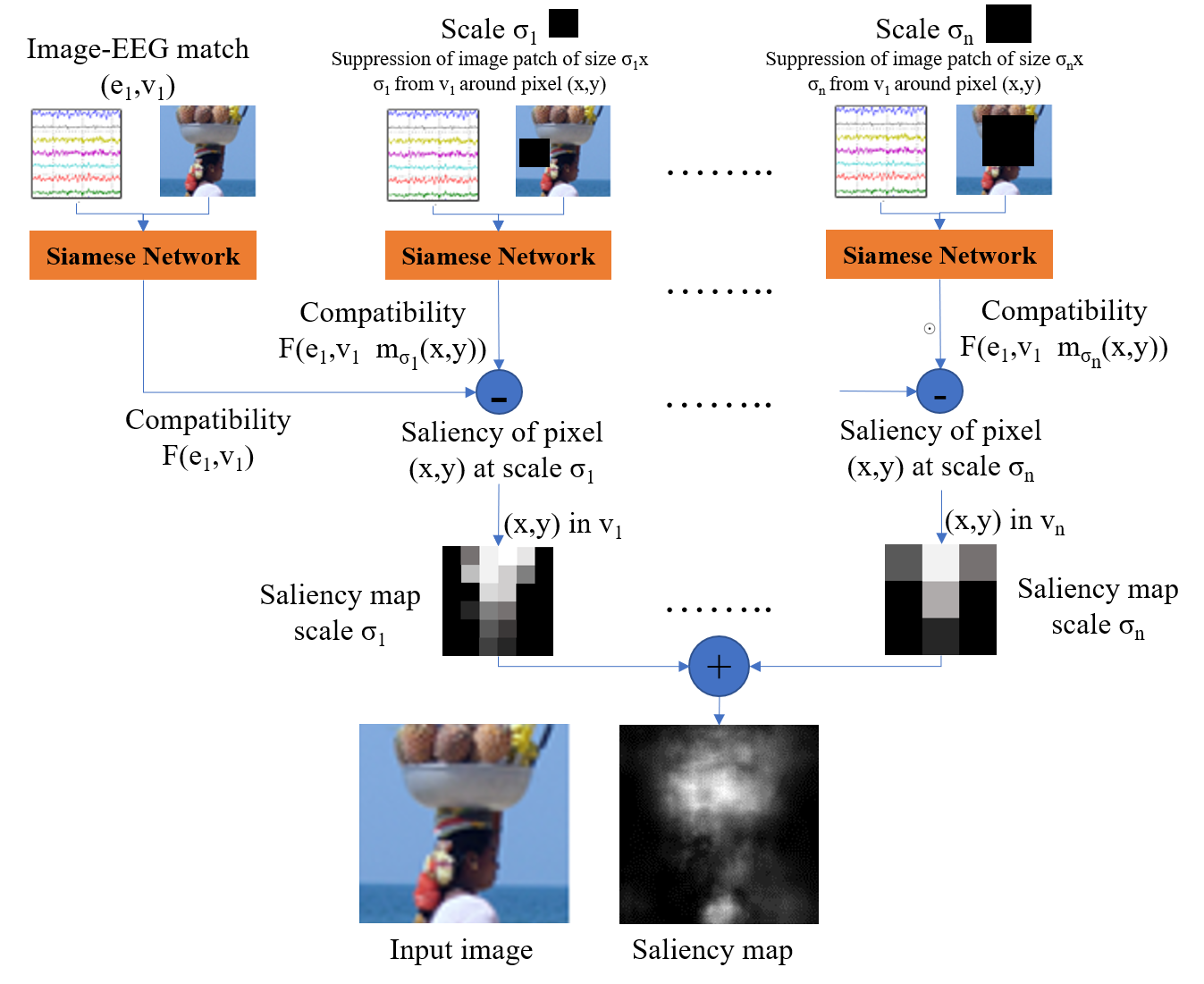}
	\caption{\textbf{Our multiscale suppression-based saliency detection}. Given an EEG/image pair, we estimate the saliency of an image patch by masking it and computing the corresponding variation in compatibility. Performing the analysis at multiple scales and for all image pixels results in a saliency map of the whole image.
	Note that, although the example scale-specific saliency maps appear pixellated, that is only a graphical artifact to give the effect of scale: in practice, scale-specific maps are still computed pixel by pixel. }
	\label{fig:saliency_method}
\end{figure}

The saliency detection method is illustrated in Fig.~\ref{fig:saliency_method} and can be formalized as follows. Let $(e,v)$ be an EEG/image pair, with compatibility $F(e,v)$. The saliency value $S(x,y, \sigma,e,v)$ at pixel  $(x,y)$ and scale $\sigma$ is obtained by removing the $\sigma \times \sigma$ image region around $(x,y)$ and computing the difference between the original compatibility score and the one after suppressing that patch. More formally, if $m_\sigma(x,y)$ is a binary mask where all pixels within the $\sigma \times \sigma$ window around $(x,y)$ are set to zero, we have:
\begin{equation}
 S(x,y,\sigma,e,v) = F(e,v) - F(e, m_\sigma(x,y) \odot v)~,
 \label{eq:sal_scale}
\end{equation}
where $\odot$ denotes element-wise multiplication (Hadamard product). For multiple scale values, we set the overall saliency value for pixel $(x, y)$ to the normalized sum of (per scale) saliency scores:
\begin{equation}
 S(x,y,e,v) = \sum_\sigma S(x,y,\sigma,e,v)~.
\end{equation}
Normalization is then performed on an image-by-image basis for visualization. 

%% file: decoding_method.tex
\section{Visual-related brain processes}\label{sec:eeg_analysis}

While the saliency detection approach studies how alterations in images reflect on compatibility scores, it is even more interesting to analyze how neural patterns act on the learned representations. Indeed, following the principle that large variations in compatibility can be found when the most important visual features are masked, we may similarly expect compatibility to drop when we remove ``important'' (from a visual feature--matching point of view) components from neural activity signals. Performing this analysis traditionally requires a combination of \emph{a priori} knowledge on brain signal patterns and manual analysis: for example, it is common to investigate the effect of provided stimuli while monitoring the emergence of event-related potentials (ERPs) known to be associated to specific brain processes. 
Of course, posing the problem in this way still requires that the processes under observation be at least partially known, which makes it complicated to automatically detect previously-unknown signal patterns.

Instead, the joint representation makes it easy to correlate brain signals with visual stimuli by analyzing how compatibility varies in response to targeted modifications of the inputs. Thus, similar to saliency detection, we can identify the spatial components in brain activity that convey visual information. 

As mentioned in Sect.~\ref{sec:related}, object recognition in humans is performed by a multi-level aggregation of shape and feature information across cortical regions, resulting in a distributed representation that can easily adapt to a wide variety of tasks on the received stimuli. For these reasons, understanding how this distributed representation is spatially localized over the brain cortex is a fundamental step towards a successful emulation of the human visual system. In order to evaluate the importance of each EEG channel (and corresponding brain area), we employ the learned joint embedding space to ``filter'' (the exact procedure is defined below) that channel from the EEG signal and measure the corresponding change in compatibility between images and filtered signals.

The importance of each channel for a single EEG/image pair can be measured by computing the difference between the pair's compatibility score and the compatibility obtained when suppressing that channel from the EEG signal. \change{Ideally, given a generic EEG/image pair $(e,v)$, and indicating with $e_{-c}$ a transformation of $e$ such that information on channel $c$ is suppressed, we could define the \emph{importance} of channel $c$ for the $(e,v)$ pair as:}
\begin{equation}
 \change{I(e, v, c) = F(e,v) - F(e_{-c}, v)~.}
 \label{eq:chan_imp}
\end{equation}
The intuition behind this formulation is that the suppression of a channel that conveys unnecessary information (at least, from the point of view of the representation learned by the EEG encoder) should result in a small difference in the compatibility score; analogously, if a channel contains important information that match brain activity data to visual data, compatibility should drop when that channel is suppressed.

\change{In practice, finding a single ideal replacement of channel $c$ to compute $e_{-c}$ is hard, since different substitutions yield varying compatibility scores. However, we found that averaging compatibility differences over a large number of random replacements of the $c$ channel gives stable results: hence, we modify Eq.~\ref{eq:chan_imp} to compute the importance score as the expected value of the difference in compatibility when replacing channel $c$ with sequences of random Gaussian samples, low-pass filtered at 100 Hz and distributed according to the original channel's estimated statistics (mean and variance).}

More formally, if EEG signal $e$ is represented as a matrix with one channel per row:
\begin{equation}
 e = \left( \begin{matrix} e_1 \\ e_2 \\ \dots \\ e_c \\ \dots \\ e_n \end{matrix} \right)~,
\end{equation}
we compute $I(e,v,c)$ as:

 \change{
\begin{equation}
\begin{split}
 I(e,v,c) = F(e,v) - \mathbb{E}\left[ F\left(\left( \begin{matrix} e_1 \\ e_2 \\ \dots \\ H\left(\mathcal{N}(\mu_c, \sigma^2_c)_{L \times 1}\right) \\ \dots \\ e_n \end{matrix}  \right), v\right) \right]~,
\end{split}
\label{eq:chan_imp_2}
\end{equation}
}
\noindent where $\mu_c$ and $\sigma^2_c$ are the sample mean and variance for channel $c$, $L$ is the EEG temporal length, $\mathcal{N}(\mu, \sigma^2)_{N \times M}$ is an $N \times M$ matrix sampled from the specified distribution, and $H$ is a low-pass filter at 100 Hz. 

\change{Finally, since channel importance scores computed over single EEG/image pairs may not be significant by themselves to draw general conclusions, we extend the definition of channel importance over multiple data samples:}
\begin{equation}
 \change{I(c) = \mathbb{E}_{(e,v)} \left[ I(e,v,c) \right]~,}
 \label{eq:chan_imp_global}
\end{equation}
\change{where the expectation is computed over all dataset samples (or a subset thereof, e.g. when grouping by class)}.

\section{Decoding brain representations}
\label{sec:decoding}

Each of the previous approaches investigated the effect of altering either the brain activity signals or the image content, but they are limited in that the differential analysis they provide is carried out in only one modality: we can identify the visual features that impact the similarity between two corresponding encodings the most, or we can identify the spatial patterns in brain activity that are more relevant to the learned representation. However, we still do not know \emph{which} visual features give rise to \emph{which} brain responses, i.e. \emph{neural generators}. To fill this gap, we propose an additional modality for interpreting compatibility differences, by employing the learned manifold to carry out an analysis of the EEG channels --- and, therefore, the corresponding brain regions --- that are most solicited in the detection of visual characteristics at different scales, from edges to textures to objects and visual concepts. To carry out this analysis, we evaluate the differences in compatibility scores computed when specific feature maps in the image encoder are removed, and map the corresponding features to the EEG channels that appear to be least active (compatibility-wise) when those features were removed. In practice, given EEG/image pair~$(e,v)$, let us define $F(e,v_{-l,f})$ as the value of the compatibility function computed by suppressing the $f$-th feature map at the $l$-th layer of the image encoder. 
According to Eq.~\ref{eq:chan_imp_2}, given EEG/image pair $(e,v)$, the importance of channel $c$ computed when a certain layer's feature is removed is:

\change{
\begin{IEEEeqnarray}{l}
I(e,v_{-l,f},c) = \nonumber \\ F(e,v_{-l,f}) - \mathbb{E}\left[ F\left(\left( \begin{matrix} e_1 \\ e_2 \\ \dots \\ H\left(\mathcal{N}(\mu_c, \sigma^2_c)_{L \times 1}\right) \\ \dots \\ e_n \end{matrix}  \right), v_{-l,f}\right) \right]~.
\end{IEEEeqnarray}
}

We then define the \emph{association} between feature $(l,f)$ and channel $c$ for a pair $(e,v)$ as follows:
\begin{equation}
 \change{A(e,v,c,l,f) = I(e,v_{-l,f},c) - I(e,v,c)}~.
\end{equation}
We consider channel $c$ and feature $(l,f)$ ``associated'' if, after removing the intrinsic importance score for that channel for a given $(e,v)$ pair, the variation in compatibility for channel $c$ does not vary when that feature is removed, which would mean no visual component in the encoded representation is left unmatched.

We can estimate the association between channel $c$ and layer $l$ by averaging over all features in that layer:
\begin{equation}
 \change{A(e,v,c,l) = \mathbb{E}_{f}\left[ A(e,v,c,l,f) \right]}~.
 \label{eq:assoc}
\end{equation}
The resulting score provides an interesting indication of how much the features computed at a certain layer in a computational model resemble the features processed by the brain in specific scalp locations.

\change{Finally, as for the channel importance score, we can compute general association scores by averaging over the entire dataset:
 \begin{equation}
  A(c,l) = \mathbb{E}_{(e,v)}\left[ A(e,v,c,l) \right]~.
  \label{eq:assoc_global}
 \end{equation}
 }

%% file: intro_experiments.tex
\change{Before we apply our joint learning strategy, we first tested the EEG classification accuracy of the EEG-ChannelNet model (described in Sect. \ref{sec:model_implement}) on the brain-visual dataset. The objective of this stage is to investigate the extent to which EEG data encodes visual information, along with the nature and importance of EEG temporal and spectral content. Lastly, we hope to provide a baseline for the  subsequent investigations. }
We then evaluate the quality and meaningfulness of the joint encoding learned by our model. The main objective is to assess the correspondence of visual and neural content in the shared representation:
\begin{itemize}
	\item \emph{Brain signal/image classification}: we evaluate if and how the learned neuro-visual manifold is representative of the two input modalities by assessing their capabilities to support classification tasks (i.e., brain signal classification, and image classification). We additionally compare the obtained results to the same models trained in a traditional supervised classification tasks. 
	\item \emph{Visual saliency detection from neural activity/image compatibility variations}: the similarity of the mapped neural signals and images should be based on the most salient features in each image, as described in Sect.~\ref{sec:saliency_method}. This evaluation, therefore, assesses the performance of our brain-based saliency detection and compares it to the state of the art methods.
	\item \emph{Localizing neural processes related to visual content}: this experiment identifies neural locations (as indicated by learned EEG scalp activity), related to specific image patches by using the method described in Sect. \ref{sec:eeg_analysis}. By combining the results of this analysis with the saliency detection results, we obtain the first ever retinotopic saliency map created by training an artificial model with salient visual features correlated with neural activity.
	\item \emph{Correlating deep learned representations with brain activity}: by analyzing the learned visual and neural patterns, we identify the most influential learned visual features  (kernels) and how they correlate with neural activity. The outcome of this evaluation indicates roughly what visual features are correlated with human visual representations. This is an important first step towards developing a methodology to better uncover and emulate human brain mechanisms.
\end{itemize}

%% file: dataset.tex
In order to test and validate our method we acquired EEG data from 6 subjects.

\CS{The recording protocol included 40 object classes with 50 images each, taken from the ImageNet dataset~\cite{imagenet_cvpr09}, giving a total of 2,000 images. Object classes were chosen according to the following criteria:
\begin{itemize}
    \item The object classes should be known and recognizable by all the subjects with a single glance;
    \item The object classes should be conceptually distinct and distant from each other (e.g., dog and cat categories are a good choice, whereas German shepherd and Dalmatian are not);
    \item The images corresponding to an object class should occupy a large portion of the image, and background of the image should be minimally complex (e.g., no other salient or distracting objects in the image).
\end{itemize}}

\CS{Visual stimuli were presented to the users in a block-based setting, with images of each class shown consecutively in a single sequence. Each image was shown for 0.5 seconds. A 10-second black screen (during which we kept recording EEG data) was presented between class blocks.}

\CS{The collected dataset contains in total 11,964 segments (time intervals recording the response to each image); 36 have been excluded from the expected 6$\times$2,000 = 12,000 segments due to low recording quality or subjects not looking at the screen, checked by using the eye movement data described in Sect.~\ref{sec:saliency}}. Each EEG segment contains 128 channels, recorded for 0.5 seconds at 1 kHz sampling rate, represented as a 128$\times$L matrix, with $L\approx 500$ being the number of samples contained in each segment on each channel. The exact duration of each signal may vary, so we discarded the first 20 samples (20 ms) to reduce interference from the previous image and then cut the signal to a common length of 440 samples (to account for signals with $L < 500$).

\begin{figure}
    \centering
    \includegraphics[width=0.45\textwidth]{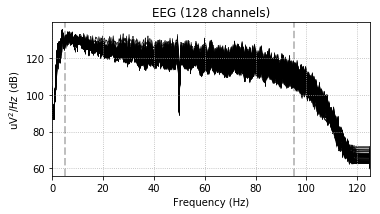}
    \caption{\change{Power spectral density of a subject's EEG recording, after 5-95 Hz band-pass filtering and notch filtering at 50 Hz.}}
    \label{fig:power}
\end{figure}

All signals were initially frequency-filtered. In detail, we applied a second-order Butterworth bandpass filter between 5 Hz and 95 Hz and a notch filter at 50 Hz. The filtered signal is then z-scored --- per channel --- to obtain zero-centered values with unitary standard deviation. We use 95 Hz as high-frequency cut-off since frequencies above \mytilde100 Hz rarely have the power to penetrate through the skull.
An example of the power spectral density of a subject's EEG recording after filtering is shown in Fig.~\ref{fig:power}. 

\begin{figure}
	\centering
	\includegraphics[width=0.24\textwidth]{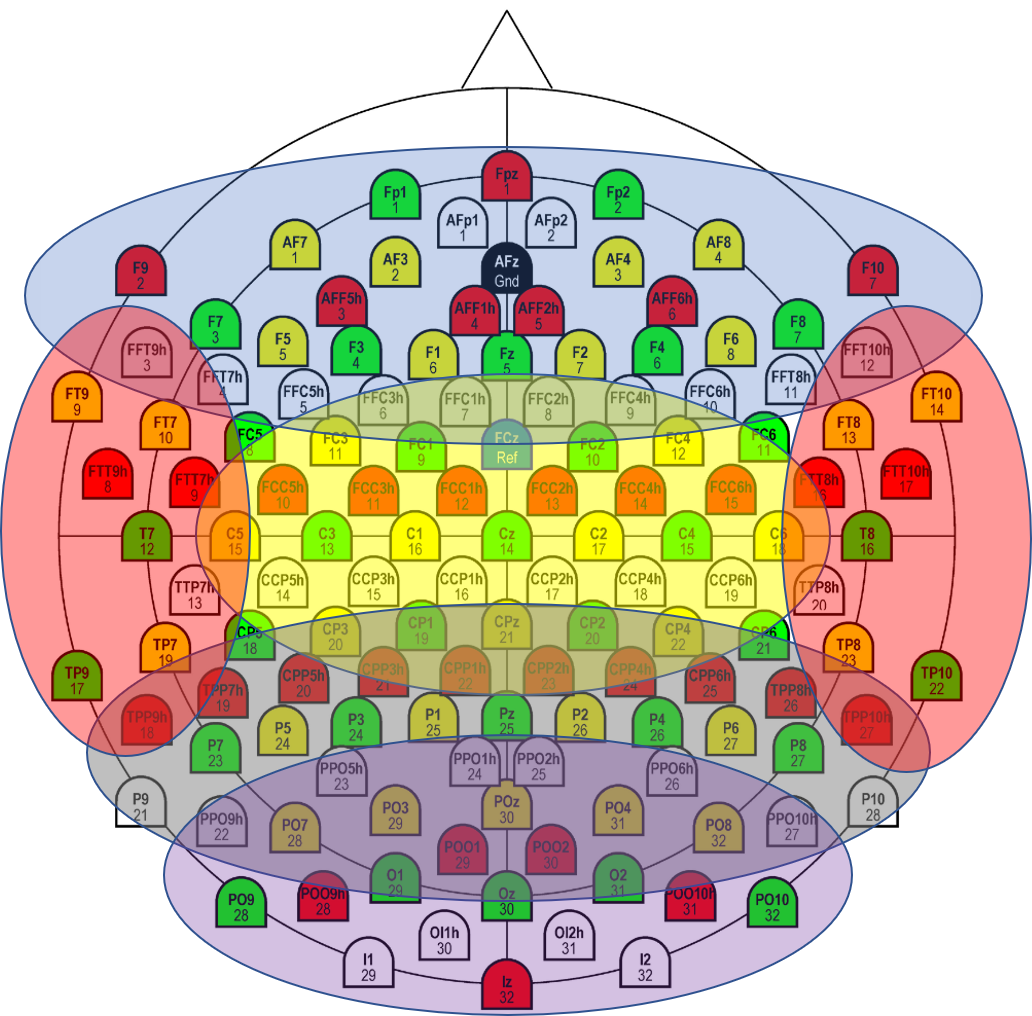} 
	\includegraphics[width=0.23\textwidth]{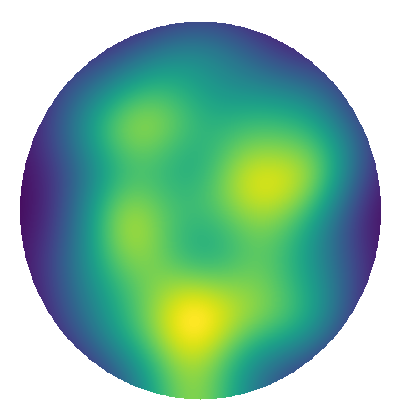}\\ \vspace{0.5cm}
	\includegraphics[width=0.48\textwidth]{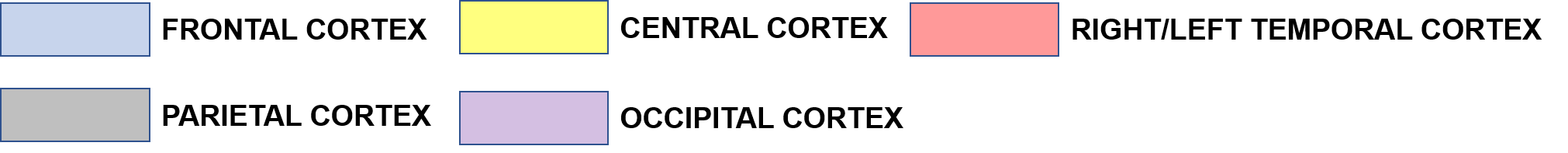}
	\caption{\textbf{Mapping between EEG channels and brain cortices.}	(Left) EEG channel placement and corresponding brain cortices (background image source: Brain Products GmbH, Gilching, Germany). We use a 128-channel EEG, where each channel in the figure is identified by a prefix letter referring to brain cortex (Fp: frontal, T: temporal, C: central, P: parietal, O: occipital) and a number indicating the electrode. (Right) Neural activation visualization --- top view of the scalp --- employed in this paper. A detailed mapping between EEG channels and brain cortices can be found in \cite{pmid11275545}.}
	\label{fig:eeg_cap}
\end{figure}

EEG placement and a mapping to the  brain cortices is shown in Fig. \ref{fig:eeg_cap}, where we also show the neural activity visualization scale employed in this paper. Activity heatmaps for an image/EEG pair are generated by applying Eq.~\ref{eq:chan_imp_2} and~\ref{eq:chan_imp_global} to estimate how much each channel affects the pair's compatibility, then plotting normalized channel importance scores on a 2D map of the scalp (at the positions corresponding to the electrodes of the employed EEG cap), and applying a Gaussian filter for smoothing (using a kernel with standard deviation of 13 pixels, for a 400$\times$400 map).

In order to replicate the training conditions employed in~\cite{Spampinato2016deep} and to make a fair comparison for brain signal classification, we use the same training, validation and test splits of the EEG dataset for the block-based design signals, consisting respectively of 1600 (80\%), 200 (10\%), 200 (10\%) images with associated EEG signals, ensuring that all signals related to a given image belong to the same split. 

%% file: eeg_encoder_impl.tex
\change{In our implementation of the EEG encoder described in Sect.~\ref{sec:encoders}, we employ five convolutional layers in the temporal block, with increasing kernel dilation values (1, 2, 4, 8, 16), and four convolutional layers in the spatial block, with kernel sizes ranging from 128 (i.e., including all EEG channels) to 16, scaling down in powers of two. Note that in both cases, the convolutional layers do not process the input in cascade, but are applied in parallel and the corresponding outputs are concatenated along the feature dimension. On the contrary, the following residual block contains four sequential residual layers, that finally lead to a non-residual convolution and a fully-connected layer that projects to the joint embedding space with a  size of 1000.}

\change{Fig.~\ref{fig:eeg_encoder} (bottom part) shows the details of the architecture in terms of layer parameters and feature map sizes. For an efficient implementation of the temporal and spatial 1D convolution, we treat the input EEG signal as a one-channel bidimensional ``image'' of size 1 (feature map) $\times$ 128 (channels) $\times$ 440 (time), and apply 2D convolutions with one dimension of the kernels equal to 1 (namely, $1\times K$ for temporal layers, and $K \times 1$ for spatial layers). 
Finally, note that padding values were used in the temporal and spatial blocks so that the output of each layer has the same size concatenation dimensions.}

%% file: classification.tex
\CS{The previous section tested the EEG-ChannelNet architecture and compared it to the state of the art standard supervised EEG classification task. This showed that our method is capable of uncovering brain activity patterns, and demonstrated the frequency bands and temporal ranges that produce the highest classification accuracy. This provides a clear baseline to understand the contribution of our methods.} In this section, we describe the training procedure for our siamese network and evaluate the quality of the learned joint embedding. In particular, we investigate a) what configurations of the two EEG and image encoders  (defined in Sect. \ref{sec:joint}) provide the best trade-off between EEG and image classification, b) how conditioning the classifier for one modality over the other affects classification accuracy, and c) if augmenting the visual representation space with features derived from the brain leads to better performance than state-of-the-art methods that only use visual features. 

We train our siamese network (the EEG and image encoders), by sampling a triplet $(e_i, v_i, v_j)$ of one EEG ($e_i$) and two images ($v_i$, $v_j$), representing, the positive $(e_i, v_i)$ and negative $(e_i, v_j)$ samples. Similar to the pure classification experiment, we used an Adam optimization algorithm with hyperparameters for our contrastive loss, a mini-batch size of 16, and the number of training epochs was set to 100.
We also test different configurations of the image encoders to investigate whether the achieved results are independent of the underlying model. 
In particular, we employ different image classification backbones as feature extractors, namely, ResNet-101, DenseNet-161, Inception-v3, and AlexNet. All of these models are first pre-trained on the ImageNet dataset, and then fine-tuned during our siamese network training. We perform data augmentation by generating multiple crops for an image associated to a given EEG sample. In particular, we resize each image by a factor of 1.1 with respect to the image encoder's expected input size (299$\times$299 for Inception-v3, 224$\times$224 for the others). We then extract ten crops from the four corners and the center of the input image, with corresponding horizontal flips. 

Once training is completed, we use the trained EEG and image encoders as feature extractors in the joint embedding space, followed by a softmax layer, for both image and EEG classification. The classification tasks provide a way to assess the quality of our multimodal learning approach and allow us to identify the best encoders' layouts, based on the accuracy of the validation set. 
The specific values for the number of convolutional layers, layer sizes, number of filters, manifold size are empirically derived to produce the best validation performance in our experiments.

\begin{table}
	\centering
	\begin{tabular}{cccc}
		\toprule
		{\textbf{Image encoder}}& \textbf{EEG} & \textbf{Image} & \textbf{Avg}\\
		\midrule
		Inception-v3  & 60.4 \% 	& 94.4 \% & 77.4 \%\\ 
		ResNet-101     	& 50.3 \% 	& 90.5 \% & 70.4 \%\\ 
		DenseNet-161   	& 54.7 \% 	& 92.1 \% & 73.4 \%\\ 
		AlexNet   	  	& 46.2 \% 	& 69.4 \% & 57.8 \%\\ 
		\bottomrule
	\end{tabular}
	\caption{EEG and image classification accuracy obtained using the joint-learning approach, for different layouts of the image encoders.} 
	\label{tab:class_base}
\end{table}

Table~\ref{tab:class_base} shows the obtained EEG and image classification accuracy for all the tested models. \CS{Note that all configurations benefit from the joint embedding learning, and achieve a classification accuracy on par or better than when training the EEG encoder alone in the standard supervised classification scenario.}

\CS{Next, we test the impact of one modality on the other, i.e., the effect of jointly learning brain activity--derived features and visual features with respect to training single-modality models. 
We first compare the image classification performance obtained by the pre-trained image encoders alone and by the image encoders obtained after fine-tuning with our joint-embedding approach. Both our model and pre-trained visual encoders are used as feature extractors, followed by a softmax layer, and performance is computed on the test split of the employed visual dataset. Note that since the 40 target images classes are included in ImageNet, the pre-trained visual encoders were previously trained on them. Therefore, we simply perform fine-tuning with the joint embedding learning, i.e., the pre-trained visual encoders are trained to maximize  the correlation between the visual and EEG content, rather than on classification \emph{per se}.}
The results in Tab.~\ref{tab:eeg_multiple_img_encoders} indicate that learning features that maximize EEG-visual correlation (as discussed in Sect. \ref{sec:joint}) leads to enhanced performance in all models. The largest increase occurred when AlexNet was the image encoder. 
This is likely due to the fact that the other models are complex enough to ``saturate'' the classification capacity (i.e. there is a ceiling effect), and suggests that the proposed approach might be useful for domain-specific tasks (e.g., medical imaging) that are particularly complex and/or where data may be limited.

\begin{table}
	\centering
	\begin{tabular}{ccc}
		\toprule
		\multicolumn{3}{c}{\textbf{Image classification performance}}\\
		\midrule
		Model &\textbf{Visual Learning} & \textbf{Joint Learning} \\
		\midrule
		Inception-v3 	& 93.1 \% &  94.4 \%  \\
		ResNet-101 		& 90.3 \% &  90.5 \%  \\
		DenseNet-161 	& 91.4 \% &  92.1 \%  \\
		AlexNet 		& 65.5 \% &  69.4 \%  \\
        \midrule
        	&  &    \\
        \midrule
		\multicolumn{3}{c}{\textbf{EEG classification performance}}\\
		\midrule

        \textbf{EEG encoder} & \textbf{EEG Learning} & \textbf{Joint Learning}\\
	    \midrule
        EEG-ChannelNet    &   48.1\% & 60.4\%\\
    \bottomrule
	\end{tabular}
	\caption{Comparison of image and EEG classification performance when using only one modality (either image or EEG) relative to when we use the joint neural-visual features. For each model, we report the best performance according to Tab.~\ref{tab:class_base}. The reported EEG classification performance for our approach are achieved when training the image encoder using Inception-v3.}
	\label{tab:eeg_multiple_img_encoders}
\end{table}

Analogously, we compare the performance of the EEG signal classification accuracy to the EEG encoder described in Sect. \ref{sec:joint}  and the one obtained by our joint neural-visual learning. The results are given in Table~\ref{tab:eeg_multiple_img_encoders} and show that the addition of visual features to the EEG classification improves performance by about 12 percentage points. Thus, the proposed joint learning scheme allows us to bring EEG classification from 48.1\%, using state-of-the-art approaches (see Tab. \ref{tab:eeg_results}), to 60.4\%. 

By comparing performance of the EEG and image classification in Tab.~\ref{tab:eeg_multiple_img_encoders}, it is important to note that the EEG classification benefits more from the use of both modalities than the image classification does. This is not surprising, given the high image only classification accuracy, and the noisy and mostly-unexplored nature of neural activity data. In this case, the integration of the more easily-classifiable visual features helps to ``guide'' the learning from the neural data to create more discriminative representations and to produce a better-performing model. Note, however, EEG classification relies on the features computed by the EEG encoder guided by the joint visual representation, rather than classifying the visual information itself (i.e. visual features are not employed during the EEG classification). Importantly, the addition of the EEG information improved performance the most when image classification was lower (i.e. classification accuracy was not yet at ceiling). This possibly suggests that when human classification accuracy is far higher than model classification accuracy, the neural data helps more.

%% file: saliency.tex
In the previous experiments, we demonstrated that the learned EEG/image embedding is able to encode enough visual information to perform both EEG and image classification. Now we investigate if and how the shared visual-brain space relates to visual saliency using the approach described in Sect.~\ref{sec:saliency_method}.  We measured how compatibility between the trained encoders and various image patches changes. The values for the $\sigma$ parameter in Eq.~\ref{eq:sal_scale} are set to 3, 5, 9, 17, 33, and 65 pixels. Note that this evaluation does not require any additional training, and can be based on the same EEG and image encoders as described in Sect.~\ref{sec:classification}. However, in order to avoid bias due to pre-trained encoders on the same dataset, the saliency experiments are carried out on re-trained versions of the models using a leave-one-out setup: for each visual class in our dataset, new EEG and image encoders are trained on the remaining 39 classes, using the same joint-embedding configuration described in Sect.~\ref{sec:classification}. \CS{In this case, compatibility essentially measures how much a given image patch accounts for the joint representation. This serves as a measure of the importance of the given image patch; patches associated to large drops in compatibility must contribute more to the joint representation.  }

For this analysis, we used eye movement data recorded --- through a 60-Hz Tobii T60 eye-tracker --- on the same six subjects of above, at the same time of EEG data acquisition, i.e., while they were looking the 2,000 displayed images. We employed this data as saliency detection dataset and the images were divided into the same training, validation and test splits of the EEG classification experiment. As a baseline comparison, we used the pre-trained SALICON~\cite{salicon} and SalNet~\cite{salnet} models, fine-tuned on the dataset's training data.\\
In addition, to demonstrate that EEG indeed encodes visual saliency information and that the generated maps are not simply driven by the image encoder, we include an additional baseline by implementing an approach similiar to the one described in Sect.~\ref{sec:saliency_method}. We used the pre-trained Inception-v3 visual classifier because it produces better classification performance (see Tab.~\ref{tab:class_base}). We then apply the same multi-scale patch-suppression method, however, in this case, the saliency score is not based on compatibility, but rather it is based on the log-likelihood variation for the image's correct class. More formally, given image $v$ and denoting  with $p(v)$ the log-likelihood of $v$'s correct class as estimated by a pre-trained Inception-v3 network, the saliency value $S_\text{classifier}(x,y, \sigma,v)$ at pixel  $(x,y)$ and scale $\sigma$ is computed as:
\begin{equation}
 S_\text{classifier}(x,y,\sigma,v) = p(v) - p\left(m_\sigma\left(x,y\right) \odot v\right),
 \label{eq:sal_classifier}
\end{equation}
where $m_\sigma(x,y) \odot v$, as previously, is the result of the removal of the $\sigma \times \sigma$ region around $(x,y)$. Also in this case, the computed saliency value at a certain location is the normalized sum over multiple scales.

\begin{figure}
\centering
\includegraphics[width=0.072\textwidth]{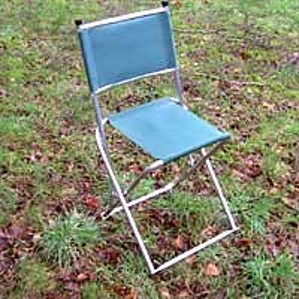}
\includegraphics[width=0.072\textwidth]{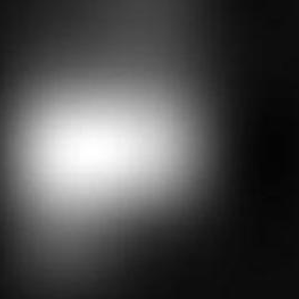}
\includegraphics[width=0.072\textwidth]{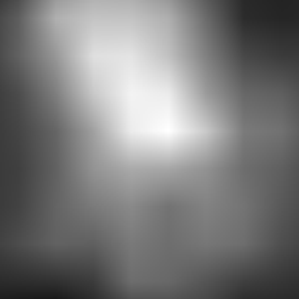}
\includegraphics[width=0.072\textwidth]{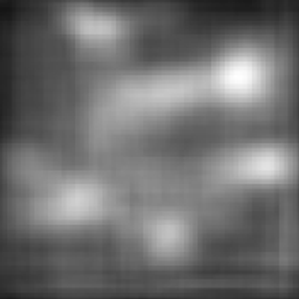}
\includegraphics[width=0.072\textwidth]{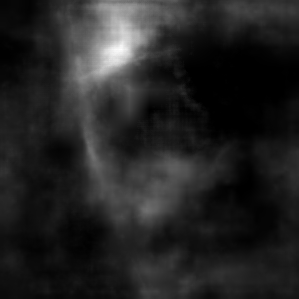}
\includegraphics[width=0.072\textwidth]{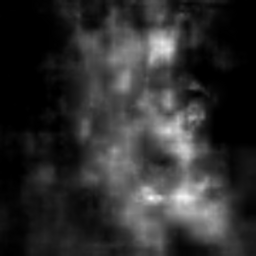}
\vspace{0.5em}\\
\includegraphics[width=0.072\textwidth]{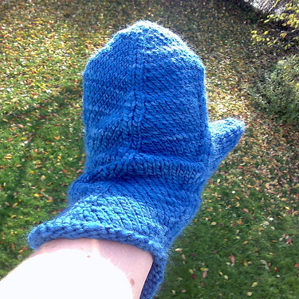}
\includegraphics[width=0.072\textwidth]{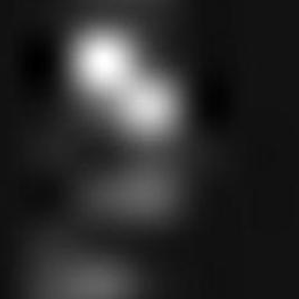}
\includegraphics[width=0.072\textwidth]{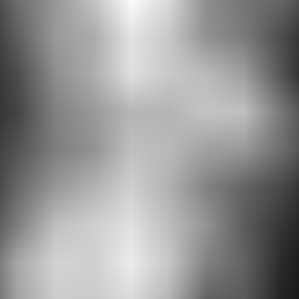}
\includegraphics[width=0.072\textwidth]{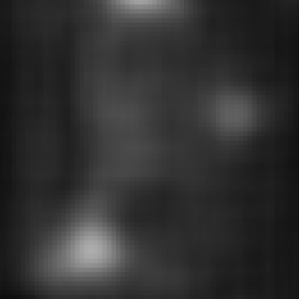}
\includegraphics[width=0.072\textwidth]{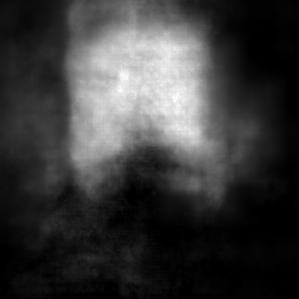}
\includegraphics[width=0.072\textwidth]{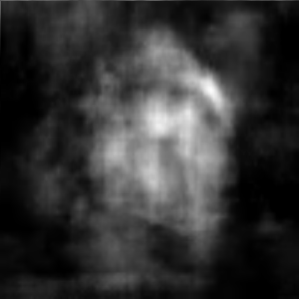}
\vspace{0.5em}\\
\includegraphics[width=0.072\textwidth]{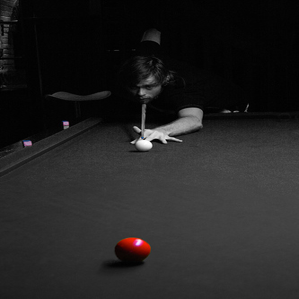}
\includegraphics[width=0.072\textwidth]{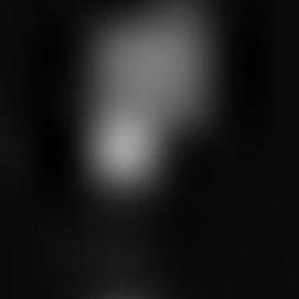}
\includegraphics[width=0.072\textwidth]{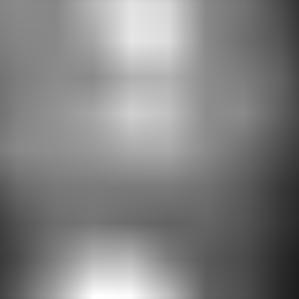}
\includegraphics[width=0.072\textwidth]{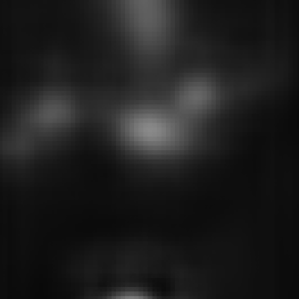}
\includegraphics[width=0.072\textwidth]{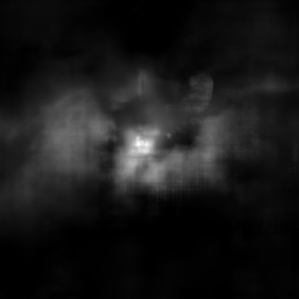}
\includegraphics[width=0.072\textwidth]{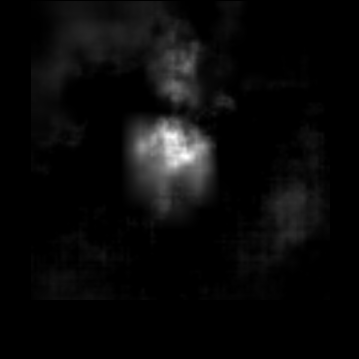}
\vspace{0.5em}\\
\includegraphics[width=0.072\textwidth]{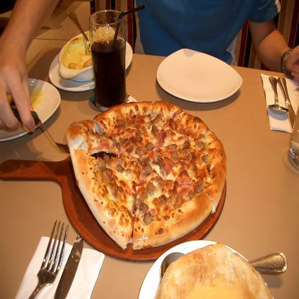}
\includegraphics[width=0.072\textwidth]{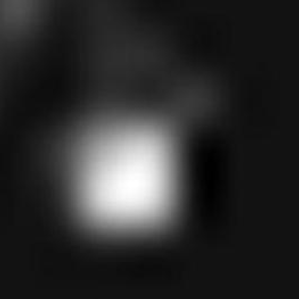}
\includegraphics[width=0.072\textwidth]{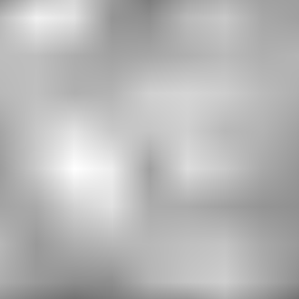}
\includegraphics[width=0.072\textwidth]{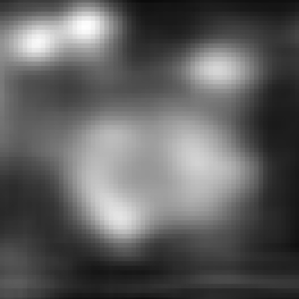}
\includegraphics[width=0.072\textwidth]{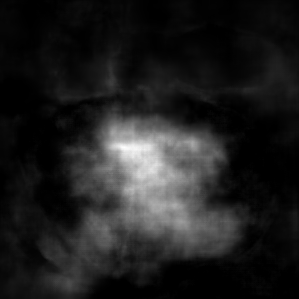}
\includegraphics[width=0.072\textwidth]{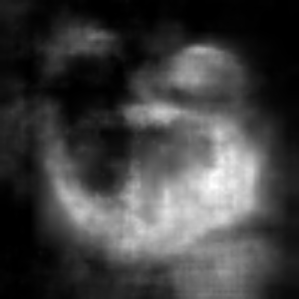}
\vspace{0.5em}\\
\includegraphics[width=0.072\textwidth]{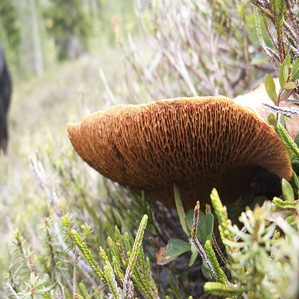}
\includegraphics[width=0.072\textwidth]{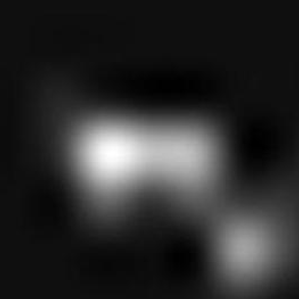}
\includegraphics[width=0.072\textwidth]{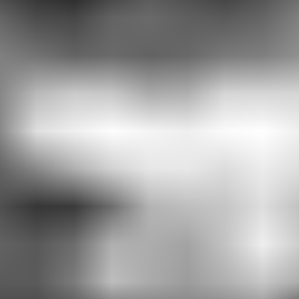}
\includegraphics[width=0.072\textwidth]{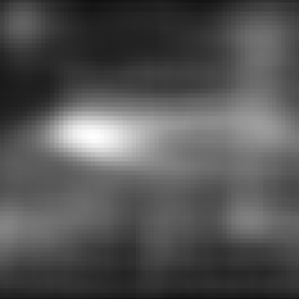}
\includegraphics[width=0.072\textwidth]{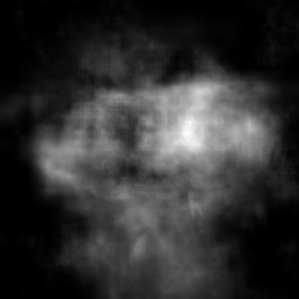}
\includegraphics[width=0.072\textwidth]{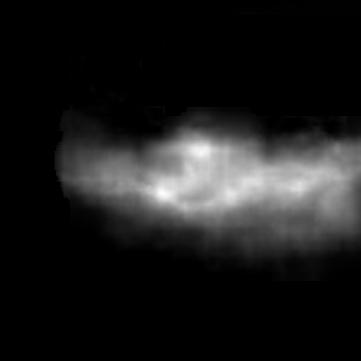}
\vspace{0.5em}\\
\caption{\change{\textbf{Qualitative comparison of generated saliency maps.} From left to right: input image, human gaze data (ground truth), SALICON, SalNet, visual classifier--driven detector, and our visual/EEG--driven detector. It can be noted a) that the maps generated by our method resemble the ground truth masks more than the state-of-the-art methods;  b) adding brain activity information to visual features results in an improved reconstruction (more details and less noise) in the saliency calcualtion (compare the 5\textsuperscript{th} and 6\textsuperscript{th} columns).}}
\label{fig:saliency_results}
\end{figure}

Fig.~\ref{fig:saliency_results} shows qualitatively the saliency maps obtained by our approach, relative to the state-of-the-art saliency detectors~\cite{salicon,salnet} and our baseline. 
We also quantitatively assess the accuracy of the maps generated by our joint-embedding--driven saliency detector by computing the metrics defined by \cite{borji2013} --- shuffled area under curve (s-AUC), normalized scanpath saliency (NSS) and correlation coefficient (CC) scores. 
\CS{Tab.~\ref{tab:saliency} reports the results achieved by our saliency detection method, showing  a) that our method outperforms the baseline saliency detectors and b) the contribution of the joint neural/visual features improving performance w.r.t. visual features alone. Importantly, this suggests that our joint embedding method accounts for more of the regions that human subjects fixate during free viewing (i.e. empirically derived saliency) than any other tested saliency methods or visual classification alone. It is also interesting to note that the metric for which our method yields the largest improvement is NSS, which is the most relevant to the nature of EEG signals, being related to the gaze fixation scan path and thus measuring a temporal aspect of saliency.}

\begin{figure}
\centering
\includegraphics[width=0.49\textwidth]{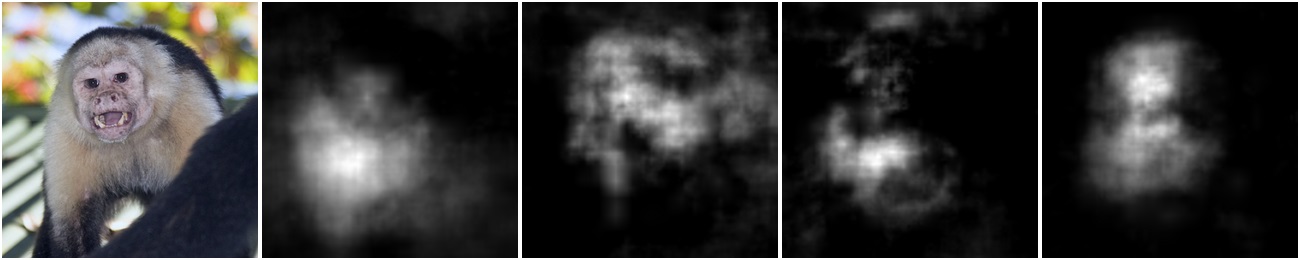}\\ \vspace{0.05cm}
\includegraphics[width=0.49\textwidth]{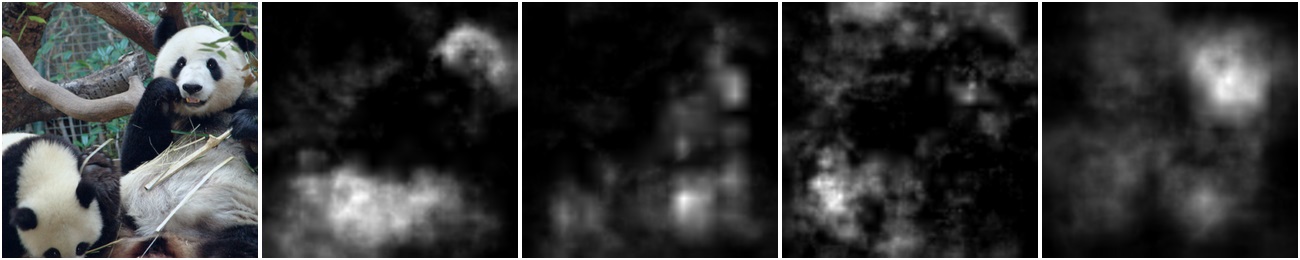}\\\vspace{0.05cm}
\includegraphics[width=0.49\textwidth]{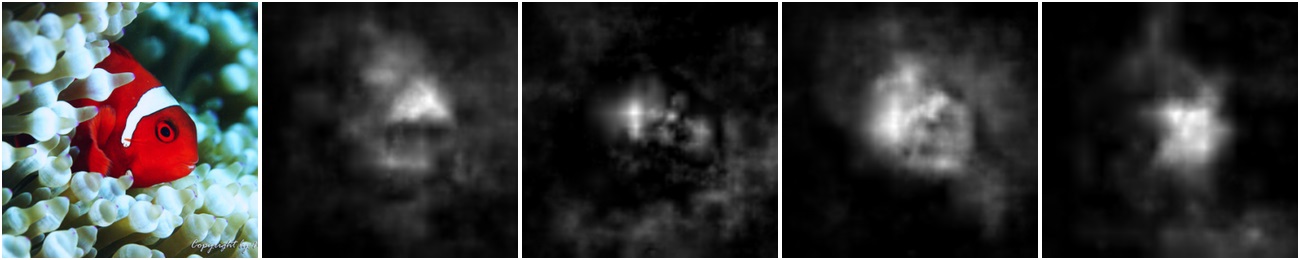}\\\vspace{0.05cm}
\includegraphics[width=0.49\textwidth]{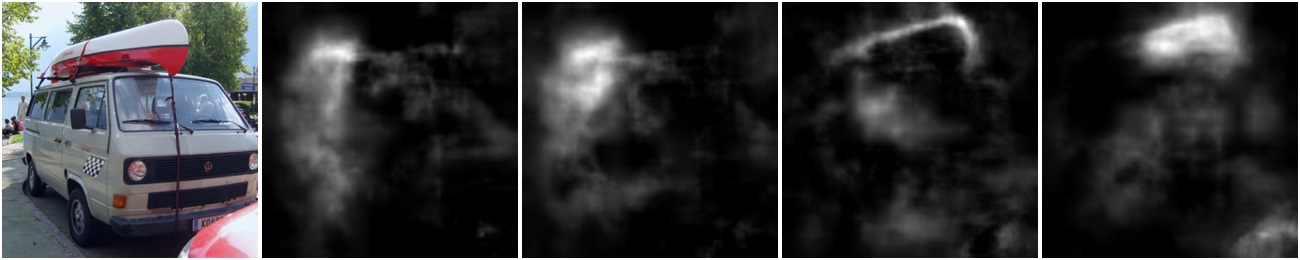}\\\vspace{0.05cm}
\includegraphics[width=0.49\textwidth]{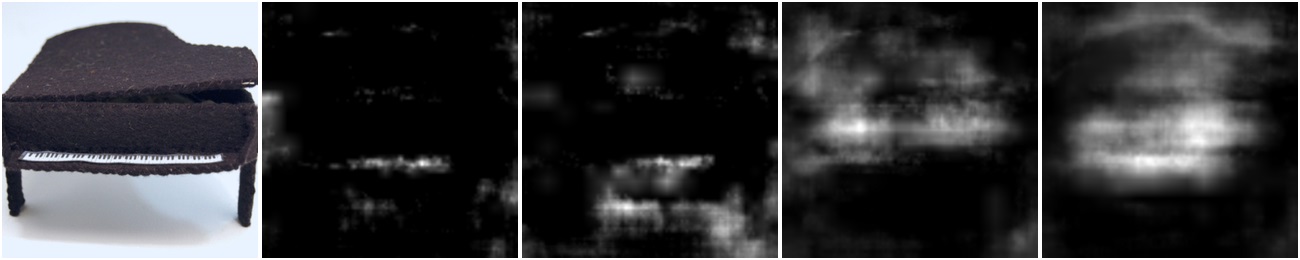}\\\vspace{0.05cm}
\includegraphics[width=0.49\textwidth]{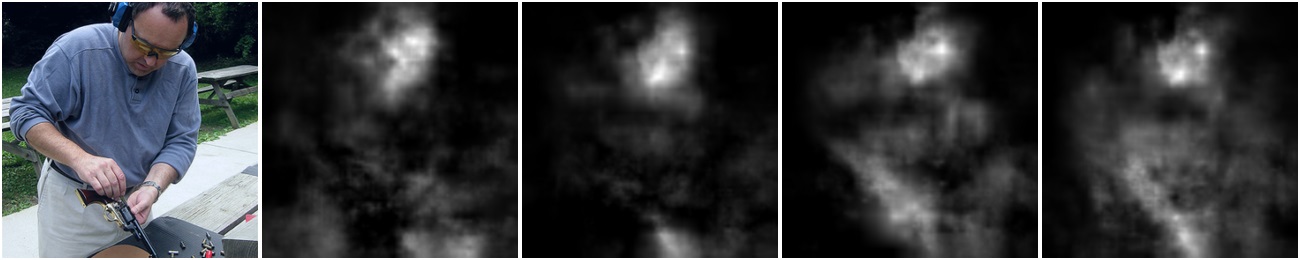}\\\vspace{0.05cm}
\caption{\textbf{Qualitative evaluation of saliency detection at different times.} From left to right: input image, saliency detection using EEG data in time range [20--240] ms, saliency detection in time range [130--350] ms, saliency detection in time range [240--460] ms and saliency detection using the entire EEG time course, i.e, [20--460] ms. It can be noted that, at the beginning, saliency is more focused on local and global visual features, and later focused on context and ultimately on objects of interest; with the last column integrating all contributions in one saliency map.}
\label{fig:saliency_subsamples}
\end{figure}

\begin{figure}
\centering
\includegraphics[width=0.49\textwidth]{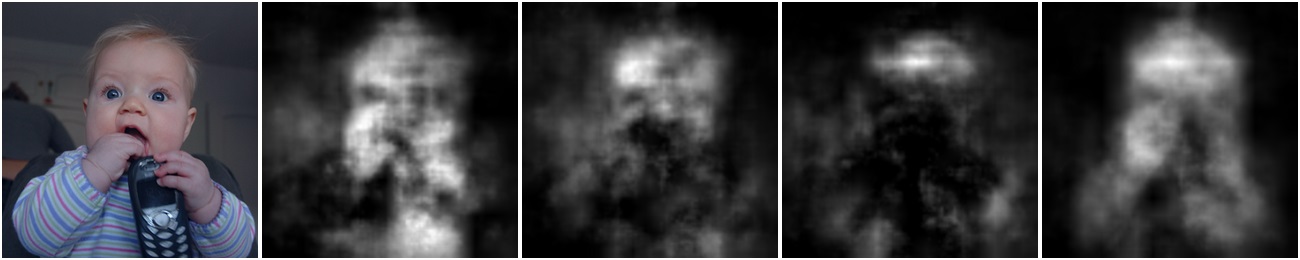}\\ \vspace{0.05cm}
\includegraphics[width=0.49\textwidth]{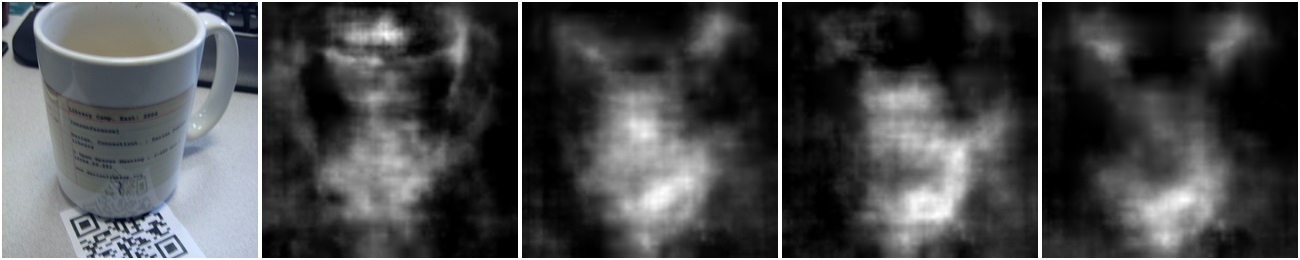}\\  \vspace{0.05cm}
\includegraphics[width=0.49\textwidth]{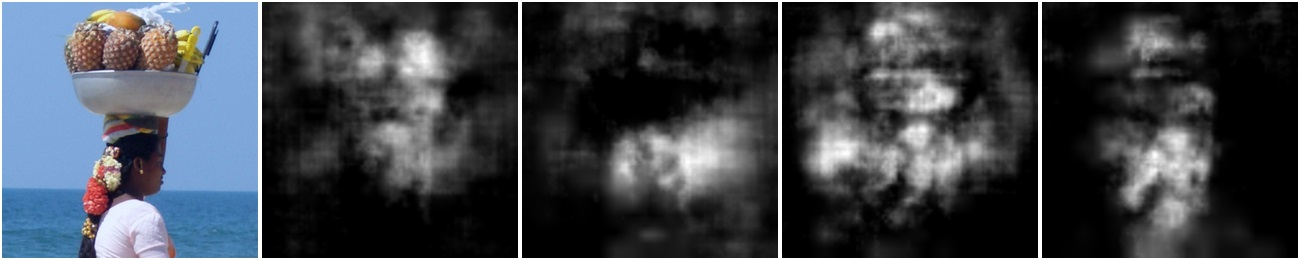}\\  \vspace{0.05cm}
\includegraphics[width=0.49\textwidth]{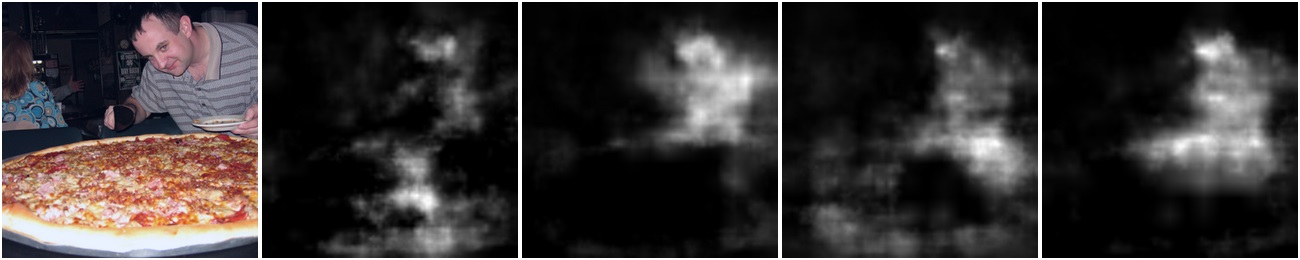}\\ 
\caption{\change{\textbf{Examples of our brain-derived saliency detection.} In all cases, the ImageNet class (from top to bottom: ``mobile phone'', ``mug'', ``banana'' and ``pizza'') is different from objects receiving more attention by the human observers. We report the saliency in the same time ranges of Fig. \ref{fig:saliency_results}. Please note that all the four images were correctly classified by the employed visual encoder, i.e, Inception-v3.}}
\label{fig:saliency_results_class}
\end{figure}

To understand how saliency evolves over time, we additionally evaluate the importance of different temporal subsamples of the EEG signal on the saliency maps. This is similar to the procedure that was done in Sect.~\ref{sec:classification_eeg}. Fig.~\ref{fig:saliency_subsamples} shows that we tested saliency across a variety of time ranges (20--240 ms, 130--350 ms and 240--460 ms). Over time, subjects appear to focus on different parts of the image. Interestingly, early on visual attention seems to be more controlled by visual features such as color contrast and edges, while later times show that attention tends to be oriented moreso towards the context or object category (i.e., the object of most interest to the observer). 
This matches theories of visual attention in humans, i.e. early on, attention is dominated by an early bottom-up unconscious process driven by basic visual features such as color, luminance, orientation, and edge detection; whereas later on, attention is driven by top-down process, which bias the observer towards regions that demonstrate context and consciously attract attention due to task demands \cite{ITTI20001489}. 
Furthermore, saliency changing over time shows that humans also pay attention to basic visual features as well as context. This is consistent with the idea that object categorization in humans is based on a combination of object and context features \cite{OLIVA2007520}.
Finally, Fig. \ref{fig:saliency_results_class} indicates that the the saliency derived using both brain and visual features is not strictly connected to the features necessary for visual recognition. For example, in the first row of Fig. \ref{fig:saliency_results_class}, the ImageNet class is ``mobile phone'' but the derived saliency focuses  more on the baby (and in the employed image dataset there is no \emph{face/person} class). This holds for all the reported examples.

\begin{table}\small
	\centering
	\begin{tabular}{clccc}
		\toprule
		& \textbf{Method} & \textbf{s-AUC} & \textbf{NSS} &  \textbf{CC} \\
		\midrule
		& SalNet    & 0.637	&	0.618	& 0.271 \\
		& SALICON 	& 0.678	& 	0.728	& 0.348\\
		& Visual classifier--driven detector& 0.532	& 0.495 &	0.173\\
		
		& \textbf{Our neural-driven detector} & \textbf{0.643}	& \textbf{0.942} & 
		\textbf{0.357}\\
		\midrule
		& Human Baseline    & 0.939&	3.042	&1 \\
		\bottomrule
	\end{tabular}\\
	\caption{Saliency performance comparison in terms of shuffled area under curve (s-AUC), normalized scanpath saliency (NSS) and correlation coefficient (CC) between our compatibility-driven saliency detector and the baseline models. We also report the human baseline, i.e., the scores computed using the ground truth maps. Since we adopt a leave-out-one setup the reported values for our approach are averaged over all the 40 experiments.}	\label{tab:saliency}
\end{table}

%% file: eeg_analysis.tex
\CS{The objective of this analysis is to approximate the spatial distribution of the cortex-level representations: indeed, while the hierarchical multi-stage architecture of the human visual pathway is known, the representations generated at each stage are  poorly understood. In these experiments, we performed a coarse analysis on the global interaction between neural activity and images, and a fine analysis on the interaction between neural activity and the deep-learned visual features. This procedure allows us to identify which neural areas (scalp regions) are the most informative. Whereas the underlying cortex cannot be precisely isolated with EEG alone, this procedure points to the temporal and spatial components of the joint representation that are sensitive to relevant to visual cues. Of course, this analysis is purely qualitative, since no ``correct'' or unequivocal answer is available. Nevertheless, we believe that it is important to verify that the generated representations are intuitively meaningful and consistent with what can be expected from the neurocognitive point of view.}

\subsubsection{Global analysis of the cortical-visual representations}

\begin{figure*}
\centering
\subfloat[Anemone]{\includegraphics[width=0.12\textwidth]{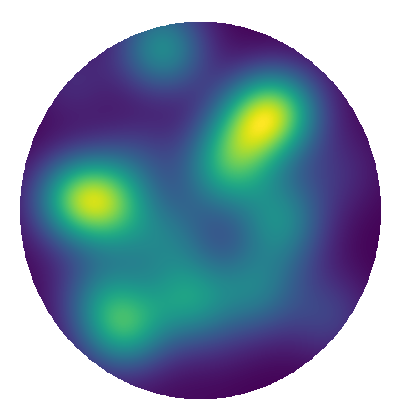}} 
\subfloat[Panda]{\includegraphics[width=0.12\textwidth]{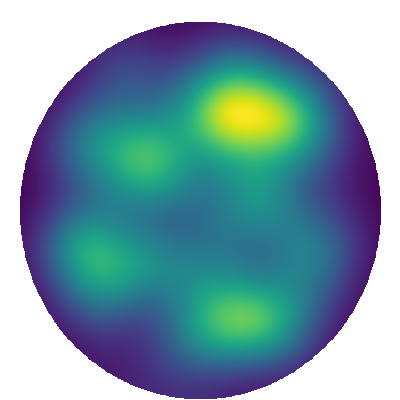}} 
\subfloat[Cellular phone]{\includegraphics[width=0.12\textwidth]{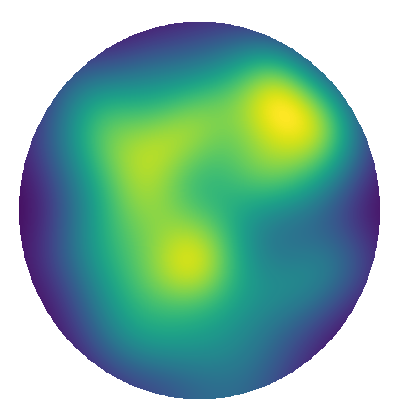}}
\subfloat[Electric guitar]{\includegraphics[width=0.12\textwidth]{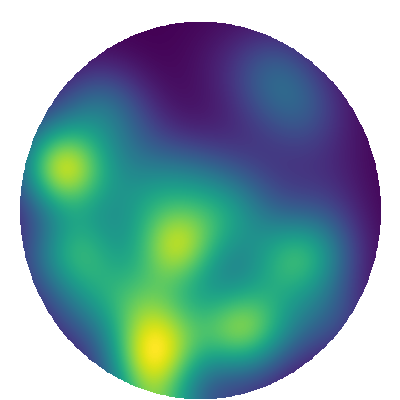}} 
\subfloat[Piano]{\includegraphics[width=0.12\textwidth]{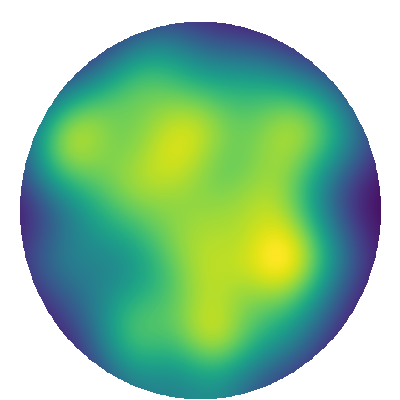}} 
\subfloat[Airliner]{\includegraphics[width=0.12\textwidth]{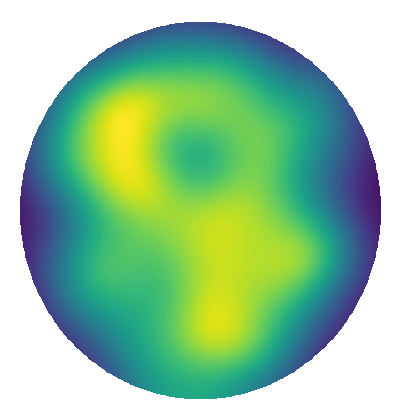}} 
\subfloat[Locomotive]{\includegraphics[width=0.12\textwidth]{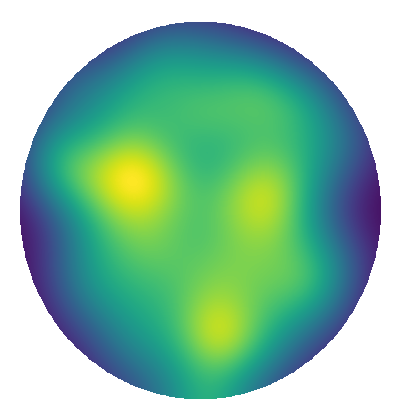}} 
\caption{\textbf{Activation maps per visual class}. Average activation maps for some of the 40 visual classes in the dataset.}
\label{fig:eeg_average_class}
\end{figure*}

\begin{figure*}
	\centering
	\subfloat[Average activation map]{ \includegraphics[width=0.15\textwidth]{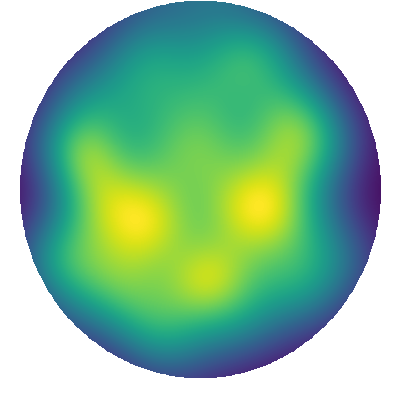} }
	\subfloat[0-80 ms]{ \includegraphics[width=0.15\textwidth]{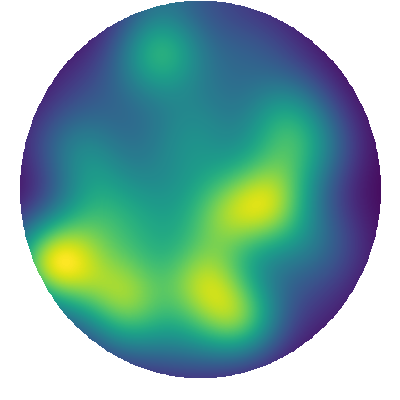} }
	\subfloat[80-160 ms]{ \includegraphics[width=0.15\textwidth]{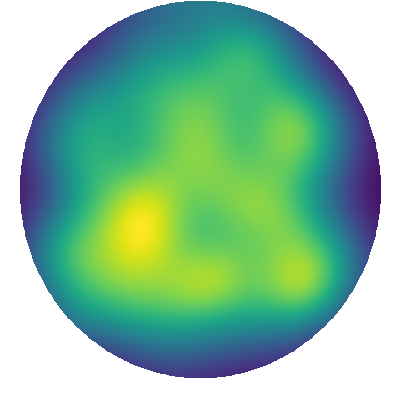} }
	\subfloat[160-320 ms]{ \includegraphics[width=0.15\textwidth]{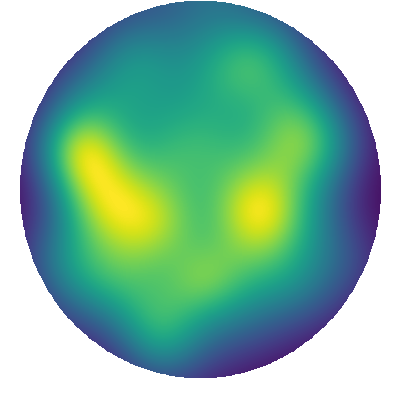} }
	\subfloat[320-440 ms]{ \includegraphics[width=0.15\textwidth]{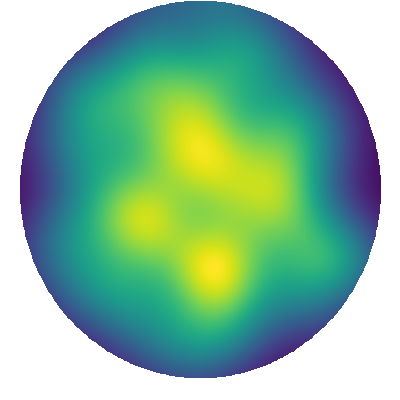} }
	\caption{\textbf{Average activation maps}. (Left image). Average activation map across all image classes. (Right images). Average activation in different time ranges.}
	\label{fig:eeg_average_intervals}
\end{figure*}

\CS{In this experiment we aim to identify high-level correlations between EEG channels and visual content, by applying Eq.~\ref{eq:chan_imp}, which assesses how average compatibility changes when each EEG channel is suppressed.} Fig.~\ref{fig:eeg_average_class} shows some examples of the mean activation maps per object class. These were obtained  by averaging channel importance scores over all images for each class. To show the relationship between the temporal and spatial activation of EEG, Fig.~\ref{fig:eeg_average_intervals} shows the average activation map over all classes, by evaluating channel importance when restricting the EEG signal to specific time intervals.

From these results, some interesting conclusions can be drawn: 1) All visual classes rely heavily on early visual areas including V1 cortex --- known to be responsible for early visual processing ~\cite{pmid29176609} --- and this region is important in all tested time windows; 2) The average activation maps over time clearly show that the process starts in early visual areas and then flows to the frontal regions (responsible of higher cognitive functions) and temporal regions (responsible for visual categorization \cite{pmid22325196}); 3) The pattern of activation changes with the visual content; e.g., the ``piano'' or the ``electric guitar''  visual class, activates scalp regions closer to auditory cortex (left-most and right-most areas of the scalp), and this is in line with evidence that the sensation of sounds is often associated with sight~\cite{pmid22355573}.

%% file: brain_representations.tex
\subsubsection{Extracting neural representations from the cortical-visual data over time}

\begin{figure*}
\centering

\subfloat[Image encoder, layer 3/20]{
 \begin{minipage}{.20\textwidth}
  \centering
  \includegraphics[width=0.23\textwidth]{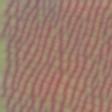}
  \includegraphics[width=0.23\textwidth]{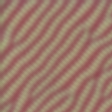}
  \includegraphics[width=0.23\textwidth]{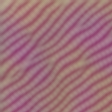}
  \includegraphics[width=0.23\textwidth]{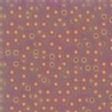}\\\vspace{0.1cm}
  \includegraphics[width=0.23\textwidth]{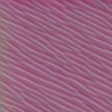}
  \includegraphics[width=0.23\textwidth]{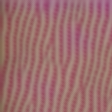}
  \includegraphics[width=0.23\textwidth]{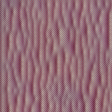}
  \includegraphics[width=0.23\textwidth]{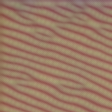}\\\vspace{0.1cm}
  \includegraphics[width=0.23\textwidth]{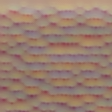}
  \includegraphics[width=0.23\textwidth]{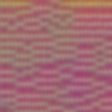}
  \includegraphics[width=0.23\textwidth]{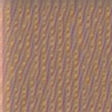}
  \includegraphics[width=0.23\textwidth]{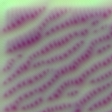}\\\vspace{0.1cm}
  \includegraphics[width=0.23\textwidth]{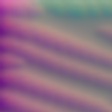}
  \includegraphics[width=0.23\textwidth]{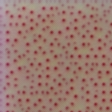}
  \includegraphics[width=0.23\textwidth]{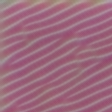}
  \includegraphics[width=0.23\textwidth]{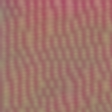}
 \end{minipage}\hspace{0.5cm}
 \begin{minipage}{.20\textwidth}
  \centering
  \includegraphics[width=1\textwidth]{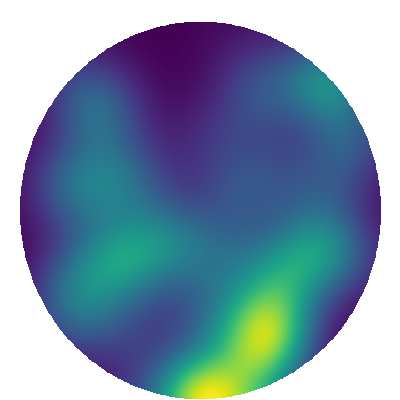}
 \end{minipage}\hspace{0.5cm}
  \begin{minipage}{.40\textwidth}
  \centering
   \includegraphics[width=0.9\textwidth, height=3.7cm]{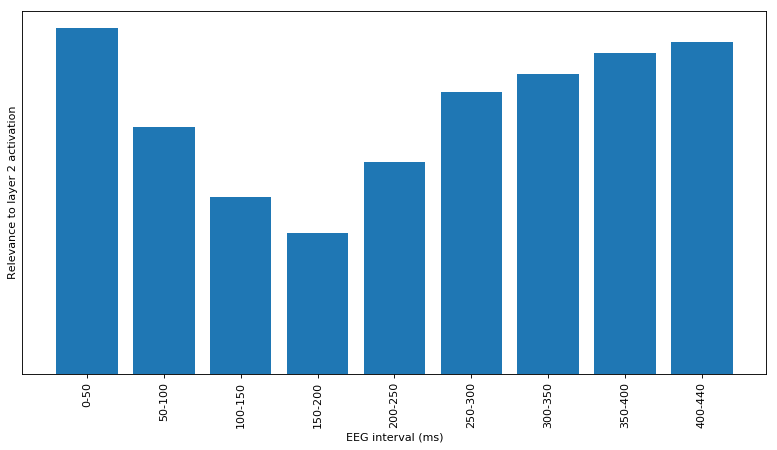}
 \end{minipage}

}

\subfloat[Image encoder, layer 12/20]{
 \begin{minipage}{.20\textwidth}
  \centering
  \includegraphics[width=0.23\textwidth]{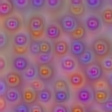}
  \includegraphics[width=0.23\textwidth]{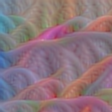}
  \includegraphics[width=0.23\textwidth]{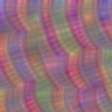}
  \includegraphics[width=0.23\textwidth]{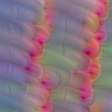}\\\vspace{0.1cm}
  \includegraphics[width=0.23\textwidth]{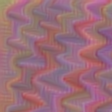}
  \includegraphics[width=0.23\textwidth]{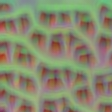}
  \includegraphics[width=0.23\textwidth]{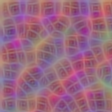}
  \includegraphics[width=0.23\textwidth]{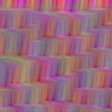}\\\vspace{0.1cm}
  \includegraphics[width=0.23\textwidth]{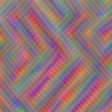}
  \includegraphics[width=0.23\textwidth]{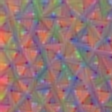}
  \includegraphics[width=0.23\textwidth]{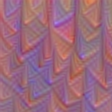}
  \includegraphics[width=0.23\textwidth]{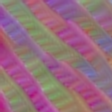}\\\vspace{0.1cm}
  \includegraphics[width=0.23\textwidth]{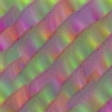}
  \includegraphics[width=0.23\textwidth]{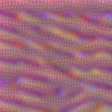}
  \includegraphics[width=0.23\textwidth]{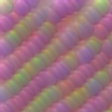}
  \includegraphics[width=0.23\textwidth]{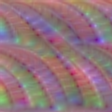}
 \end{minipage}\hspace{0.5cm}
 \begin{minipage}{.20\textwidth}
  \centering
  \includegraphics[width=1\textwidth]{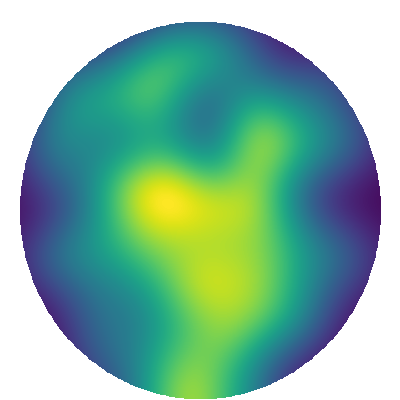}
 \end{minipage}\hspace{0.5cm}
  \begin{minipage}{.40\textwidth}
  \centering
   \includegraphics[width=0.9\textwidth, height=3.7cm]{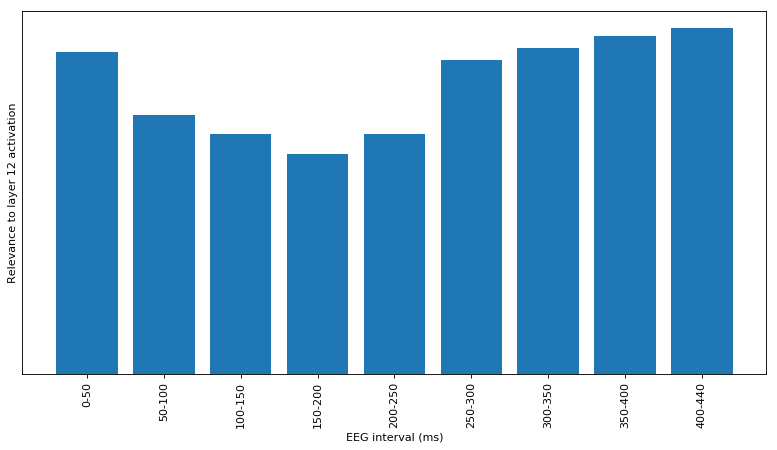}
 \end{minipage}

}

\subfloat[Image encoder, layer 20/20]{
 \begin{minipage}{.20\textwidth}
  \centering
  \includegraphics[width=0.23\textwidth]{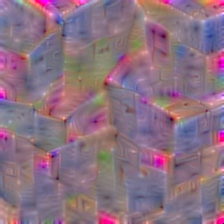}
  \includegraphics[width=0.23\textwidth]{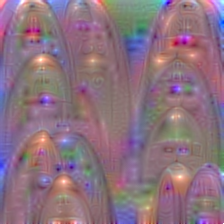}
  \includegraphics[width=0.23\textwidth]{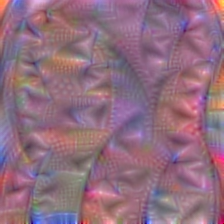}
  \includegraphics[width=0.23\textwidth]{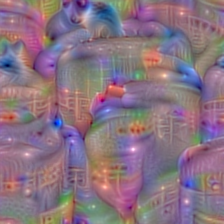}\\\vspace{0.1cm}
  \includegraphics[width=0.23\textwidth]{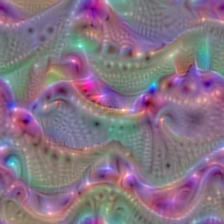}
  \includegraphics[width=0.23\textwidth]{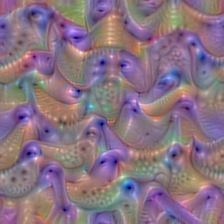}
  \includegraphics[width=0.23\textwidth]{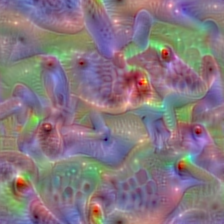}
  \includegraphics[width=0.23\textwidth]{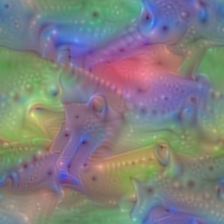}\\\vspace{0.1cm}
  \includegraphics[width=0.23\textwidth]{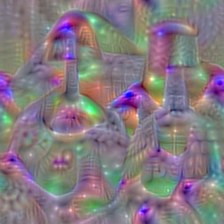}
  \includegraphics[width=0.23\textwidth]{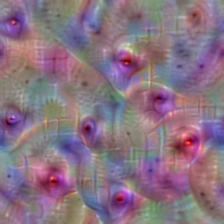}
  \includegraphics[width=0.23\textwidth]{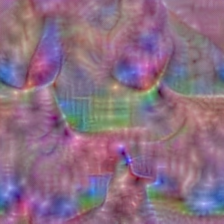}
  \includegraphics[width=0.23\textwidth]{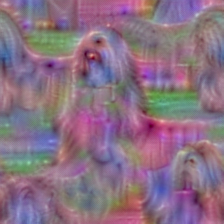}\\\vspace{0.1cm}
  \includegraphics[width=0.23\textwidth]{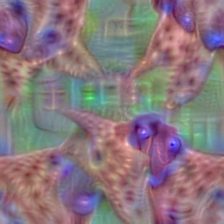}
  \includegraphics[width=0.23\textwidth]{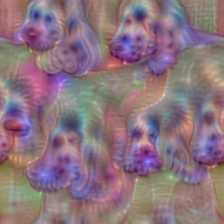}
  \includegraphics[width=0.23\textwidth]{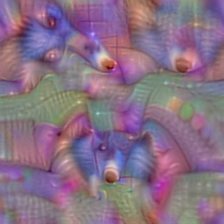}
  \includegraphics[width=0.23\textwidth]{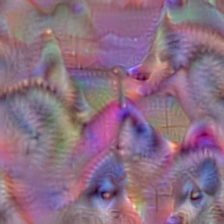}
 \end{minipage}\hspace{0.5cm}
 \begin{minipage}{.20\textwidth}
  \centering
  \includegraphics[width=1\textwidth]{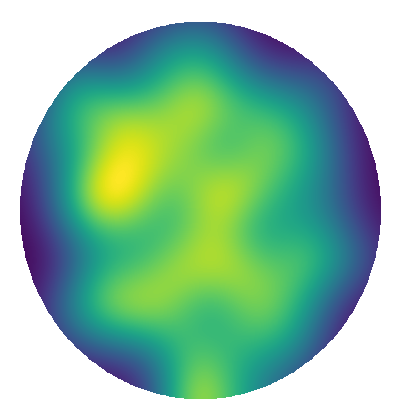}
 \end{minipage}\hspace{0.5cm}
  \begin{minipage}{.40\textwidth}
  \centering
   \includegraphics[width=0.9\textwidth, height=3.7cm]{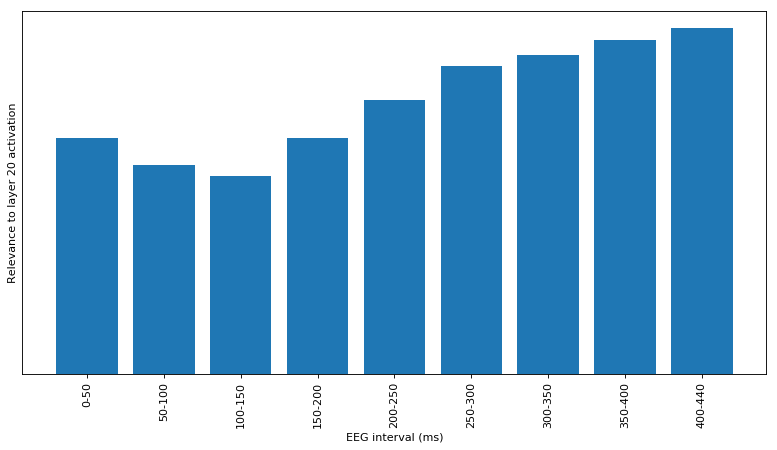}
 \end{minipage}

}

\caption{\textbf{Brain activity associated with specific visual representations extracted from the DCNN layers}. Each row shows a set of feature maps (manually picked for interpretability and visualized through activation maximization) from a specific layer in the image encoder, the neural activity areas with the highest association to the layer's features, and the contribution that different time ranges in the EEG signal give rise to association scores. It can be noted that, as feature complexity increases, the activated brain regions move from the V1 visual cortex (occipital region) to the IT cortex (temporal region); moreover, the initial temporal portions of EEG signals seem to be more related to simpler features, while there is a stronger association between more complex features and later temporal dynamics.}
\label{fig:eeg_vis_repr}
\end{figure*}

\CS{The goal of this final analysis is to probe the various DCNN layers and relate them to the joint EEG/visual data, over time and over scalp location, to examine the low- and middle-level visual representations responsible for the given neural activation.}
To accomplish this task, we employ the learned compatibility measure to find mutual correspondence between the deep features and the scalp region generating the activity. Using the \emph{association} score defined in Sect.~\ref{sec:decoding} (Eq.~\ref{eq:assoc_global}), we investigate the neural encoding of the visual information by deriving neural activation maps that maximally respond to the deep-learned visual features. 
Fig.~\ref{fig:eeg_vis_repr} shows the activation maps of the association scores related to specific layers of our best-performing image encoder as per Tab.~\ref{tab:class_base}. This analysis employs a pre-trained Inception network fine-tuned on our brain/image dataset during encoder training. To show the complexity of the features learned at each level, we show a few examples obtained by performing activation maximization~\cite{olah2017feature} on a subset of features for each layer.
For each feature/neural association, we also measure the relative contribution to brain activity by different temporal portions of the EEG, by feeding each interval to the EEG encoder (as described in the previous section) when applying Eq.~\ref{eq:assoc_global}. In this case, unlike the representations in Fig.~\ref{fig:eeg_average_intervals}, we are not interested in the differences in activation between cortical regions. \CS{Therefore, we compute the average unnormalized association scores over all channels, and use that as the measure of how associated each layer's features are with each portion of the EEG activation. 
By using all of this information, we are able to probe the underlying neural representations, the spatial location on the scalp that relate to the representation and their timing. 
The results suggest that hierarchical representations in DCNNs tightly correlate with the hierarchical processing stages in the human visual pathway. In particular, at the lowest layer, simple texture and color features are generated and they correspond with early visual areas near V1. Moving to deeper layers in the DCNN, we see that the activation propagates from early visual areas to temporal regions and then back to the early visual regions. Moreover, more complex features (at higher layers) are influenced by the activity occurring later in time. Whereas early visual areas, known to encode basic visual features, correspond to the early DCNN layers, which also encode simple visual features, later layers, which produce more complex class-level representations, seem to correspond to later EEG time windows. The timing of the EEG activity and the associated DCNN layers are in line with the known hierarchical object processing stream in the cognitive neuroscience literature. This consistency suggests that we have produced reliable approximations of human brain representations. It is interesting to note that we observe a consistent drop in the relationship between the joint EEG activation and DCNN layers in the 100-200 ms time window. Importantly, the end of this time window corresponds to the well-established transition from (primarily) perceptual processing to (primarily) higher order, cognitive and recurrent processing~\cite{luck2014introduction}. This suggests a logical relationship to known human neural processing. Alternatively, this could originate from the relocation of visual cognitive processes to deeper cortical areas that are less detectable via EEG, followed by feedback activity to the initial regions in the visual pathway. Clearly, future work will need to explore these interesting possibilities further, since a comprehensive neurological interpretation is outside of the scope of this paper.}

%% file: conclusions.tex
In this work, we present a multi-modal approach to learn a joint feature space for images and EEG signals recorded while users look at pictures on a screen. We trained two encoders in a siamese configuration and maximize the \emph{compatibility} score between the corresponding images and EEGs. The learned embeddings make the representation useful to perform several computer vision tasks, supervised by brain activity. Our experiments show the neural activity can be reliably used to drive the development of image classification and saliency detection methods. In addition to advancing the work related to brain-guided image classification \cite{Spampinato2016deep}, our approach provides a way to extract neural representation from EEG data and to map it to the most important/salient visual features.\\
While drawing general cognitive neuroscience conclusions from these findings is not the main goal of this work, given also the small scale of the cognitive experiment, we propose an AI-based strategy that seems to produce reliable approximations of brain representations and their corresponding scalp activity, by jointly learning a model that maximizes the correlation between neural activity and visual images. \\
The natural extension of this work in the future is to further investigate these associations, with the objective of finding a finer correspondence between EEG signals and visual patterns --- e.g., by identifying different responses in the brain activity corresponding to specific objects, patterns, or categories of varying specificity. We believe that a joint research effort combining artificial intelligence (through the development of more sophisticated methods) and neuroscience (through more tailored and large scale experiments) is necessary to advance both fields, by studying how brain processes relate to artificial model structures and, in turn, using the uncovered neural dynamics to propose novel neural architectures to make computational models more closely approximate human perceptual and cognitive performance.

%% file: main.bbl
\begin{thebibliography}{10}
\providecommand{\url}[1]{#1}
\csname url@samestyle\endcsname
\providecommand{\newblock}{\relax}
\providecommand{\bibinfo}[2]{#2}
\providecommand{\BIBentrySTDinterwordspacing}{\spaceskip=0pt\relax}
\providecommand{\BIBentryALTinterwordstretchfactor}{4}
\providecommand{\BIBentryALTinterwordspacing}{\spaceskip=\fontdimen2\font plus
\BIBentryALTinterwordstretchfactor\fontdimen3\font minus
  \fontdimen4\font\relax}
\providecommand{\BIBforeignlanguage}[2]{{%
\expandafter\ifx\csname l@#1\endcsname\relax
\typeout{** WARNING: IEEEtran.bst: No hyphenation pattern has been}%
\typeout{** loaded for the language `#1'. Using the pattern for}%
\typeout{** the default language instead.}%
\else
\language=\csname l@#1\endcsname
\fi
#2}}
\providecommand{\BIBdecl}{\relax}
\BIBdecl

\bibitem{pmid28530228}
T.~Horikawa and Y.~Kamitani, ``{{G}eneric decoding of seen and imagined objects
  using hierarchical visual features},'' \emph{Nat Commun}, vol.~8, p. 15037,
  May 2017.

\bibitem{pmid27282108}
R.~M. Cichy, A.~Khosla, D.~Pantazis, A.~Torralba, and A.~Oliva, ``{{C}omparison
  of deep neural networks to spatio-temporal cortical dynamics of human visual
  object recognition reveals hierarchical correspondence},'' \emph{Sci Rep},
  vol.~6, p. 27755, 06 2016.

\bibitem{Spampinato2016deep}
C.~Spampinato, S.~Palazzo, I.~Kavasidis, D.~Giordano, N.~Souly, and M.~Shah,
  ``{Deep Learning Human Mind for Automated Visual Classification},'' in
  \emph{CVPR}, jul 2017, pp. 4503--4511.

\bibitem{8237631}
S.~Palazzo, C.~Spampinato, I.~Kavasidis, D.~Giordano, and M.~Shah, ``Generative
  adversarial networks conditioned by brain signals,'' in \emph{ICCV}, Oct
  2017, pp. 3430--3438.

\bibitem{pmid21945275}
S.~Nishimoto, A.~T. Vu, T.~Naselaris, Y.~Benjamini, B.~Yu, and J.~L. Gallant,
  ``{{R}econstructing visual experiences from brain activity evoked by natural
  movies},'' \emph{Curr. Biol.}, vol.~21, no.~19, pp. 1641--1646, Oct 2011.

\bibitem{pmid23932491}
D.~E. Stansbury, T.~Naselaris, and J.~L. Gallant, ``{{N}atural scene statistics
  account for the representation of scene categories in human visual cortex},''
  \emph{Neuron}, vol.~79, no.~5, pp. 1025--1034, Sep 2013.

\bibitem{pmid11690606}
J.~Bullier, ``{{I}ntegrated model of visual processing},'' \emph{Brain Res.
  Brain Res. Rev.}, vol.~36, no. 2-3, pp. 96--107, Oct 2001.

\bibitem{pmid21438683}
Z.~Kourtzi and C.~E. Connor, ``{{N}eural representations for object perception:
  structure, category, and adaptive coding},'' \emph{Annu. Rev. Neurosci.},
  vol.~34, pp. 45--67, 2011.

\bibitem{pmid21415848}
D.~J. Kravitz, K.~S. Saleem, C.~I. Baker, and M.~Mishkin, ``{{A} new neural
  framework for visuospatial processing},'' \emph{Nat. Rev. Neurosci.},
  vol.~12, no.~4, pp. 217--230, Apr 2011.

\bibitem{pmid22325196}
J.~J. DiCarlo, D.~Zoccolan, and N.~C. Rust, ``{{H}ow does the brain solve
  visual object recognition?}'' \emph{Neuron}, vol.~73, no.~3, pp. 415--434,
  Feb 2012.

\bibitem{pmlr-v80-wen18a}
H.~Wen, K.~Han, J.~Shi, Y.~Zhang, E.~Culurciello, and Z.~Liu, ``Deep predictive
  coding network for object recognition,'' in \emph{35th International
  Conference on Machine Learning}, ser. Proceedings of Machine Learning
  Research, J.~Dy and A.~Krause, Eds., vol.~80.\hskip 1em plus 0.5em minus
  0.4em\relax Stockholmsmässan, Stockholm Sweden: PMLR, 10--15 Jul 2018, pp.
  5266--5275.

\bibitem{pmid23663408}
A.~Clark, ``{{W}hatever next? {P}redictive brains, situated agents, and the
  future of cognitive science},'' \emph{Behav Brain Sci}, vol.~36, no.~3, pp.
  181--204, Jun 2013.

\bibitem{pmid23177956}
A.~M. Bastos, W.~M. Usrey, R.~A. Adams, G.~R. Mangun, P.~Fries, and K.~J.
  Friston, ``{{C}anonical microcircuits for predictive coding},''
  \emph{Neuron}, vol.~76, no.~4, pp. 695--711, Nov 2012.

\bibitem{roy_review_eeg}
Y.~Roy, H.~J. Banville, I.~Albuquerque, A.~Gramfort, T.~H. Falk, and
  J.~Faubert, ``Deep learning-based electroencephalography analysis: a
  systematic review,'' \emph{CoRR}, vol. abs/1901.05498, 2019.

\bibitem{AAAI1816107}
D.~Zhang, L.~Yao, X.~Zhang, S.~Wang, W.~Chen, R.~Boots, and B.~Benatallah,
  ``Cascade and parallel convolutional recurrent neural networks on eeg-based
  intention recognition for brain computer interface,'' 2018.

\bibitem{2607-18}
V.~A. N. P. A.~M. K.R.Rao, ``Cognitive analysis of working memory load from
  eeg, by a deep recurrent neural network,'' 2018.

\bibitem{YanS19.004}
P.~Yan, F.~Wang, and Z.~Grinspan, ``: Spectrographic seizure detection using
  deep learning with convolutional neural networks (s19.004),''
  \emph{Neurology}, vol.~90, no. 15 Supplement, 2018.

\bibitem{7727334}
L.~{Vidyaratne}, A.~{Glandon}, M.~{Alam}, and K.~M. {Iftekharuddin}, ``Deep
  recurrent neural network for seizure detection,'' in \emph{2016 International
  Joint Conference on Neural Networks (IJCNN)}, 2016, pp. 1202--1207.

\bibitem{8275511}
T.~{Zhang}, W.~{Zheng}, Z.~{Cui}, Y.~{Zong}, and Y.~{Li}, ``Spatial–temporal
  recurrent neural network for emotion recognition,'' \emph{IEEE Transactions
  on Cybernetics}, vol.~49, no.~3, pp. 839--847, 2019.

\bibitem{TANG201711}
Z.~Tang, C.~Li, and S.~Sun, ``Single-trial eeg classification of motor imagery
  using deep convolutional neural networks,'' \emph{Optik}, vol. 130, pp. 11 --
  18, 2017.

\bibitem{Lawhern_2018}
V.~J. Lawhern, A.~J. Solon, N.~R. Waytowich, S.~M. Gordon, C.~P. Hung, and
  B.~J. Lance, ``{EEGNet}: a compact convolutional neural network for
  {EEG}-based brain{\textendash}computer interfaces,'' \emph{Journal of Neural
  Engineering}, vol.~15, no.~5, p. 056013, jul 2018.

\bibitem{NIPS2017_7048}
\BIBentryALTinterwordspacing
Y.~Li, m.~Murias, s.~Major, g.~Dawson, K.~Dzirasa, L.~Carin, and D.~E. Carlson,
  ``Targeting eeg/lfp synchrony with neural nets,'' in \emph{Advances in Neural
  Information Processing Systems 30}, I.~Guyon, U.~V. Luxburg, S.~Bengio,
  H.~Wallach, R.~Fergus, S.~Vishwanathan, and R.~Garnett, Eds.\hskip 1em plus
  0.5em minus 0.4em\relax Curran Associates, Inc., 2017, pp. 4620--4630.
  [Online]. Available:
  \url{http://papers.nips.cc/paper/7048-targeting-eeglfp-synchrony-with-neural-nets.pdf}
\BIBentrySTDinterwordspacing

\bibitem{pmid25681421}
K.~J. Seymour, M.~A. Williams, and A.~N. Rich, ``{{T}he {R}epresentation of
  {C}olor across the {H}uman {V}isual {C}ortex: {D}istinguishing {C}hromatic
  {S}ignals {C}ontributing to {O}bject {F}orm {V}ersus {S}urface {C}olor},''
  \emph{Cereb. Cortex}, vol.~26, no.~5, pp. 1997--2005, May 2016.

\bibitem{pmid26053241}
J.~W. Peirce, ``{{U}nderstanding mid-level representations in visual
  processing},'' \emph{J Vis}, vol.~15, no.~7, p.~5, 2015.

\bibitem{pmid16272124}
C.~P. Hung, G.~Kreiman, T.~Poggio, and J.~J. DiCarlo, ``{{F}ast readout of
  object identity from macaque inferior temporal cortex},'' \emph{Science},
  vol. 310, no. 5749, pp. 863--866, Nov 2005.

\bibitem{pmid29176609}
A.~K. Robinson, P.~Venkatesh, M.~J. Boring, M.~J. Tarr, P.~Grover, and
  M.~Behrmann, ``{{V}ery high density {E}{E}{G} elucidates spatiotemporal
  aspects of early visual processing},'' \emph{Sci Rep}, vol.~7, no.~1, p.
  16248, Nov 2017.

\bibitem{NIPS2013_4991}
D.~L. Yamins, H.~Hong, C.~Cadieu, and J.~J. DiCarlo, ``Hierarchical modular
  optimization of convolutional networks achieves representations similar to
  macaque it and human ventral stream,'' in \emph{Advances in Neural
  Information Processing Systems 26}, C.~J.~C. Burges, L.~Bottou, M.~Welling,
  Z.~Ghahramani, and K.~Q. Weinberger, Eds.\hskip 1em plus 0.5em minus
  0.4em\relax Curran Associates, Inc., 2013, pp. 3093--3101.

\bibitem{pmid24812127}
D.~L. Yamins, H.~Hong, C.~F. Cadieu, E.~A. Solomon, D.~Seibert, and J.~J.
  DiCarlo, ``{{P}erformance-optimized hierarchical models predict neural
  responses in higher visual cortex},'' \emph{Proc. Natl. Acad. Sci. U.S.A.},
  vol. 111, no.~23, pp. 8619--8624, Jun 2014.

\bibitem{pmid19104670}
N.~Kriegeskorte, M.~Mur, and P.~Bandettini, ``{{R}epresentational similarity
  analysis - connecting the branches of systems neuroscience},'' \emph{Front
  Syst Neurosci}, vol.~2, p.~4, 2008.

\bibitem{pmid27732574}
A.~Graves, G.~Wayne, M.~Reynolds, T.~Harley, I.~Danihelka,
  A.~Grabska-Barwi?ska, S.~G. Colmenarejo, E.~Grefenstette, T.~Ramalho,
  J.~Agapiou, A.~P. Badia, K.~M. Hermann, Y.~Zwols, G.~Ostrovski, A.~Cain,
  H.~King, C.~Summerfield, P.~Blunsom, K.~Kavukcuoglu, and D.~Hassabis,
  ``{{H}ybrid computing using a neural network with dynamic external memory},''
  \emph{Nature}, vol. 538, no. 7626, pp. 471--476, 10 2016.

\bibitem{pmlr-v37-gregor15}
K.~Gregor, I.~Danihelka, A.~Graves, D.~Rezende, and D.~Wierstra, ``Draw: A
  recurrent neural network for image generation,'' in \emph{32nd International
  Conference on Machine Learning}, ser. Proceedings of Machine Learning
  Research, F.~Bach and D.~Blei, Eds., vol.~37.\hskip 1em plus 0.5em minus
  0.4em\relax Lille, France: PMLR, 07--09 Jul 2015, pp. 1462--1471.

\bibitem{pmlr-v37-xuc15}
K.~Xu, J.~Ba, R.~Kiros, K.~Cho, A.~Courville, R.~Salakhudinov, R.~Zemel, and
  Y.~Bengio, ``Show, attend and tell: Neural image caption generation with
  visual attention,'' in \emph{32nd International Conference on Machine
  Learning}, ser. Proceedings of Machine Learning Research, F.~Bach and
  D.~Blei, Eds., vol.~37.\hskip 1em plus 0.5em minus 0.4em\relax Lille, France:
  PMLR, 07--09 Jul 2015, pp. 2048--2057.

\bibitem{pmid27881212}
B.~M. Lake, T.~D. Ullman, J.~B. Tenenbaum, and S.~J. Gershman, ``{{B}uilding
  machines that learn and think like people},'' \emph{Behav Brain Sci},
  vol.~40, p. e253, Jan 2017.

\bibitem{pmid29599461}
R.~C. Fong, W.~J. Scheirer, and D.~D. Cox, ``{{U}sing human brain activity to
  guide machine learning},'' \emph{Sci Rep}, vol.~8, no.~1, p. 5397, Mar 2018.

\bibitem{4587618}
A.~Kapoor, P.~Shenoy, and D.~Tan, ``Combining brain computer interfaces with
  vision for object categorization,'' in \emph{2008 CVPR}, June 2008, pp. 1--8.

\bibitem{pmid26353347}
W.~J. Scheirer, S.~E. Anthony, K.~Nakayama, and D.~D. Cox, ``{{P}erceptual
  {A}nnotation: {M}easuring {H}uman {V}ision to {I}mprove {C}omputer
  {V}ision},'' \emph{IEEE Trans Pattern Anal Mach Intell}, vol.~36, no.~8, pp.
  1679--1686, Aug 2014.

\bibitem{conf/icml/NgiamKKNLN11}
J.~Ngiam, A.~Khosla, M.~Kim, J.~Nam, H.~Lee, and A.~Y. Ng, ``Multimodal deep
  learning.'' in \emph{ICML}, L.~Getoor and T.~Scheffer, Eds.\hskip 1em plus
  0.5em minus 0.4em\relax Omnipress, 2011, pp. 689--696.

\bibitem{NIPS2014_5279}
K.~Sohn, W.~Shang, and H.~Lee, ``Improved multimodal deep learning with
  variation of information,'' in \emph{Advances in Neural Information
  Processing Systems 27}, Z.~Ghahramani, M.~Welling, C.~Cortes, N.~D. Lawrence,
  and K.~Q. Weinberger, Eds.\hskip 1em plus 0.5em minus 0.4em\relax Curran
  Associates, Inc., 2014, pp. 2141--2149.

\bibitem{JMLR:v15:srivastava14b}
N.~Srivastava and R.~Salakhutdinov, ``Multimodal learning with deep boltzmann
  machines,'' \emph{Journal of Machine Learning Research}, vol.~15, pp.
  2949--2980, 2014.

\bibitem{Venugopalan_2017_CVPR}
S.~Venugopalan, L.~Anne~Hendricks, M.~Rohrbach, R.~Mooney, T.~Darrell, and
  K.~Saenko, ``Captioning images with diverse objects,'' in \emph{The CVPR
  (CVPR)}, July 2017.

\bibitem{NIPS2017_6658}
I.~Ilievski and J.~Feng, ``Multimodal learning and reasoning for visual
  question answering,'' in \emph{Advances in Neural Information Processing
  Systems 30}, I.~Guyon, U.~V. Luxburg, S.~Bengio, H.~Wallach, R.~Fergus,
  S.~Vishwanathan, and R.~Garnett, Eds.\hskip 1em plus 0.5em minus 0.4em\relax
  Curran Associates, Inc., 2017, pp. 551--562.

\bibitem{5540120}
M.~Guillaumin, J.~Verbeek, and C.~Schmid, ``Multimodal semi-supervised learning
  for image classification,'' in \emph{2010 IEEE Computer Society Conference on
  Computer Vision and Pattern Recognition}, June 2010, pp. 902--909.

\bibitem{zhao2018sound}
H.~Zhao, C.~Gan, A.~Rouditchenko, C.~Vondrick, J.~McDermott, and A.~Torralba,
  ``The sound of pixels,'' \emph{arXiv preprint arXiv:1804.03160}, 2018.

\bibitem{10.1007/978}
A.~Owens, J.~Wu, J.~H. McDermott, W.~T. Freeman, and A.~Torralba, ``Ambient
  sound provides supervision for visual learning,'' in \emph{Computer Vision --
  ECCV 2016}, B.~Leibe, J.~Matas, N.~Sebe, and M.~Welling, Eds.\hskip 1em plus
  0.5em minus 0.4em\relax Cham: Springer International Publishing, 2016, pp.
  801--816.

\bibitem{Aytar:20166}
Y.~Aytar, C.~Vondrick, and A.~Torralba, ``Soundnet: Learning sound
  representations from unlabeled video,'' in \emph{30th International
  Conference on Neural Information Processing Systems}, ser. NIPS'16.\hskip 1em
  plus 0.5em minus 0.4em\relax USA: Curran Associates Inc., 2016, pp. 892--900.

\bibitem{7952132}
S.~Hershey, S.~Chaudhuri, D.~P.~W. Ellis, J.~F. Gemmeke, A.~Jansen, R.~C.
  Moore, M.~Plakal, D.~Platt, R.~A. Saurous, B.~Seybold, M.~Slaney, R.~J.
  Weiss, and K.~Wilson, ``Cnn architectures for large-scale audio
  classification,'' in \emph{2017 IEEE International Conference on Acoustics,
  Speech and Signal Processing (ICASSP)}, March 2017, pp. 131--135.

\bibitem{Huiskes:2010}
M.~J. Huiskes, B.~Thomee, and M.~S. Lew, ``New trends and ideas in visual
  concept detection: The mir flickr retrieval evaluation initiative,'' in
  \emph{International Conference on Multimedia Information Retrieval}, ser. MIR
  '10.\hskip 1em plus 0.5em minus 0.4em\relax New York, NY, USA: ACM, 2010, pp.
  527--536.

\bibitem{Arandjelovic_2017_ICCV}
R.~Arandjelovic and A.~Zisserman, ``Look, listen and learn,'' in \emph{The IEEE
  International Conference on Computer Vision (ICCV)}, Oct 2017.

\bibitem{43274}
O.~Vinyals, A.~Toshev, S.~Bengio, and D.~Erhan, ``Show and tell: A neural image
  caption generator,'' in \emph{Computer Vision and Pattern Recognition}, 2015.

\bibitem{KarpathyF17}
A.~Karpathy and L.~Fei{-}Fei, ``Deep visual-semantic alignments for generating
  image descriptions,'' \emph{{IEEE} Trans. Pattern Anal. Mach. Intell.},
  vol.~39, no.~4, pp. 664--676, 2017.

\bibitem{DonahueHGRVDS15}
J.~Donahue, L.~A. Hendricks, S.~Guadarrama, M.~Rohrbach, S.~Venugopalan,
  T.~Darrell, and K.~Saenko, ``Long-term recurrent convolutional networks for
  visual recognition and description.'' in \emph{CVPR}.\hskip 1em plus 0.5em
  minus 0.4em\relax IEEE Computer Society, 2015, pp. 2625--2634.

\bibitem{pmlr-v48-reed16}
S.~Reed, Z.~Akata, X.~Yan, L.~Logeswaran, B.~Schiele, and H.~Lee, ``Generative
  adversarial text to image synthesis,'' in \emph{33rd International Conference
  on Machine Learning}, ser. Proceedings of Machine Learning Research, M.~F.
  Balcan and K.~Q. Weinberger, Eds., vol.~48.\hskip 1em plus 0.5em minus
  0.4em\relax New York, New York, USA: PMLR, 20--22 Jun 2016, pp. 1060--1069.

\bibitem{MansimovPBS15}
E.~Mansimov, E.~Parisotto, L.~J. Ba, and R.~Salakhutdinov, ``Generating images
  from captions with attention,'' \emph{ICLR2016}, vol. abs/1511.02793, 2016.

\bibitem{Kavasidis:2017}
I.~Kavasidis, S.~Palazzo, C.~Spampinato, D.~Giordano, and M.~Shah,
  ``Brain2image: Converting brain signals into images,'' in \emph{ACM MM '17},
  2017, pp. 1809--1817.

\bibitem{YuK15}
F.~Yu and V.~Koltun, ``Multi-scale context aggregation by dilated
  convolutions,'' \emph{ICLR2016}, vol. abs/1511.07122, 2015.

\bibitem{He2015deep}
K.~He, X.~Zhang, S.~Ren, and J.~Sun, ``{Deep Residual Learning for Image
  Recognition},'' in \emph{CVPR (CVPR)}, Las Vegas, USA, 2016.

\bibitem{TREUE2003428}
S.~Treue, ``Visual attention: the where, what, how and why of saliency,''
  \emph{Current Opinion in Neurobiology}, vol.~13, no.~4, pp. 428 -- 432, 2003.

\bibitem{zeiler2014visualizing}
M.~D. Zeiler and R.~Fergus, ``Visualizing and understanding convolutional
  networks,'' in \emph{European conference on computer vision}.\hskip 1em plus
  0.5em minus 0.4em\relax Springer, 2014, pp. 818--833.

\bibitem{imagenet_cvpr09}
J.~Deng, W.~Dong, R.~Socher, L.-J. Li, K.~Li, and L.~Fei-Fei, ``{ImageNet: A
  Large-Scale Hierarchical Image Database},'' in \emph{CVPR09}, 2009.

\bibitem{pmid11275545}
R.~Oostenveld and P.~Praamstra, ``{{T}he five percent electrode system for
  high-resolution {E}{E}{G} and {E}{R}{P} measurements},'' \emph{Clin
  Neurophysiol}, vol. 112, no.~4, pp. 713--719, Apr 2001.

\bibitem{CLAYTON2015188}
M.~S. Clayton, N.~Yeung, and R.~C. Kadosh, ``The roles of cortical oscillations
  in sustained attention,'' \emph{Trends in Cognitive Sciences}, vol.~19,
  no.~4, pp. 188 -- 195, 2015.

\bibitem{TALLONBAUDRY1999151}
C.~Tallon-Baudry and O.~Bertrand, ``Oscillatory gamma activity in humans and
  its role in object representation,'' \emph{Trends in Cognitive Sciences},
  vol.~3, no.~4, pp. 151 -- 162, 1999.

\bibitem{JENSEN2007317}
O.~Jensen, J.~Kaiser, and J.-P. Lachaux, ``Human gamma-frequency oscillations
  associated with attention and memory,'' \emph{Trends in Neurosciences},
  vol.~30, no.~7, pp. 317 -- 324, 2007.

\bibitem{luck2014introduction}
S.~J. Luck, \emph{An introduction to the event-related potential
  technique}.\hskip 1em plus 0.5em minus 0.4em\relax MIT press, 2014.

\bibitem{salicon}
X.~Huang, C.~Shen, X.~Boix, and Q.~Zhao, ``Salicon: Reducing the semantic gap
  in saliency prediction by adapting deep neural networks,'' in \emph{ICCV
  2015}, 2015, pp. 262--270.

\bibitem{salnet}
J.~Pan, E.~Sayrol, X.~Giro-i Nieto, K.~McGuinness, and N.~E. O'Connor,
  ``Shallow and deep convolutional networks for saliency prediction,'' in
  \emph{CVPR 2016}, 2016.

\bibitem{borji2013}
A.~Borji, D.~N. Sihite, and L.~Itti, ``Quantitative analysis of human-model
  agreement in visual saliency modeling: A comparative study,'' \emph{TIP
  2013}, 2013.

\bibitem{ITTI20001489}
L.~Itti and C.~Koch, ``A saliency-based search mechanism for overt and covert
  shifts of visual attention,'' \emph{Vision Research}, vol.~40, no.~10, pp.
  1489 -- 1506, 2000.

\bibitem{OLIVA2007520}
A.~Oliva and A.~Torralba, ``The role of context in object recognition,''
  \emph{Trends in Cognitive Sciences}, vol.~11, no.~12, pp. 520 -- 527, 2007.

\bibitem{pmid22355573}
A.~M. Proverbio, G.~E. D'Aniello, R.~Adorni, and A.~Zani, ``{{W}hen a
  photograph can be heard: vision activates the auditory cortex within 110
  ms},'' \emph{Sci Rep}, vol.~1, p.~54, 2011.

\bibitem{olah2017feature}
C.~Olah, A.~Mordvintsev, and L.~Schubert, ``Feature visualization,''
  \emph{Distill}, vol.~2, no.~11, p.~e7, 2017.

\end{thebibliography}
